\def\year{2021}\relax
\documentclass[letterpaper]{article}
\usepackage{aaai21-preprint}
\usepackage{times}
\usepackage{helvet}
\usepackage{courier}
\usepackage[hyphens]{url}
\usepackage{graphicx}
\usepackage{natbib}
\usepackage{caption}
\usepackage{epsfig}
\usepackage{amsmath}
\usepackage{amssymb}
\usepackage{comment}

\usepackage{tikz,pgfplots}

\usepackage[utf8]{inputenc}
\usepackage[T1]{fontenc}
\usepackage{url}
\usepackage{booktabs}
\usepackage{amsfonts}
\usepackage{nicefrac}
\usepackage{microtype}
\usepackage{bm,amsfonts,mathrsfs}

\newcommand{\eg}{\textit{e.g.}}
\newcommand{\ie}{\textit{i.e.}}
\newcommand{\cf}{\textit{cf.}}

\newcommand{\mathbold}[1]{\bm{#1}}
\newcommand{\mbf}[1]{\mathbf{#1}}

\newcommand{\T}{^\mathsf{T}}

\newcommand{\R}{\mathbb{R}}

\newcommand{\vtheta}[0]{\mathbold{\theta}}

\newcommand{\MLambda}[0]{\mathbold{\Lambda}}

\renewcommand{\mid}{\,|\,}

\newcommand{\vf}{\mbf{f}}

\newcommand{\vp}{\mbf{p}}
\newcommand{\vq}{\mbf{q}}

\newcommand{\vu}{\mbf{u}}

\newcommand{\vx}{\mbf{x}}
\newcommand{\vy}{\mbf{y}}
\newcommand{\vz}{\mbf{z}}

\newcommand{\MI}{\mbf{I}}

\newcommand{\MK}{\mbf{K}}

\newcommand{\MR}{\mbf{R}}

  \makeatletter
  \let\MYcaption\@makecaption
  \makeatother

  \usepackage{subcaption}

  \makeatletter
  \let\@makecaption\MYcaption
  \makeatother

\usepackage{tikz,pgfplots}
\usetikzlibrary{plotmarks}
\usetikzlibrary{decorations.text}

\pgfplotsset{/pgf/number format/.cd, 1000 sep={}}

\pgfplotsset{every axis/.append style={
  grid style={line width=0.6pt,dotted,gray}}}

\pgfplotsset{every axis/.append style={
  legend style={inner xsep=1pt, inner ysep=0.5pt, nodes={inner sep=1pt, text depth=0.1em},draw=none,fill=none}
}}

\pgfplotsset{every axis/.append style={
  colorbar style={width=3mm,xshift=-2mm,major tick length=2pt}
}}

\usetikzlibrary{external}
\tikzsetexternalprefix{external/}
\newlength{\figurewidth}
\newlength{\figureheight}

\usepackage{algorithm}
\definecolor{cgray}{gray}{0.4}

\renewcommand{\paragraph}[1]{\medskip\noindent {\bf #1}~~}

\usepackage[capitalize, nameinlink]{cleveref}
\crefname{section}{Sec.}{Secs.}

\title{Gaussian Process Priors for View-Aware Inference}
\author{
  Yuxin Hou\textsuperscript{\rm 1,*},
  Ari Heljakka\textsuperscript{\rm 1,2,*}, and 
  Arno Solin\textsuperscript{\rm 1} \\
}
\affiliations{

    \textsuperscript{\rm 1}Aalto University, Espoo, Finland \qquad
    \textsuperscript{\rm 2}GenMind Ltd., Finland \qquad
    \textsuperscript{\rm *}Equal contribution \\

    \{yuxin.hou, ari.heljakka, arno.solin\}@aalto.fi

}

\begin{document}

\maketitle

\begin{abstract}
While frame-independent predictions with deep neural networks have become the prominent solutions to many computer vision tasks, the potential benefits of utilizing correlations between frames have received less attention. Even though probabilistic machine learning provides the ability to encode correlation as prior knowledge for inference, there is a tangible gap between the theory and practice of applying probabilistic methods to modern vision problems. For this, we derive a principled framework to combine information coupling between camera poses (translation and orientation) with deep models. We proposed a novel view kernel that generalizes the standard periodic kernel in $\mathrm{SO}(3)$. We show how this soft-prior knowledge can aid several pose-related vision tasks like novel view synthesis and predict arbitrary points in the latent space of generative models, pointing towards a range of new applications for inter-frame reasoning.

\end{abstract}

\section{Introduction}
\label{sec:intro}
Gaussian processes \citep[GPs,][]{Rasmussen+Williams:2006} provide a flexible probabilistic framework for combining {\it a~priori} knowledge with forecasting, noise removal, and explaining data. Their strengths are in many ways complementary to those of deep neural networks which perform best in applications where large training data sets are available and the test points reside close to the training samples. The tremendous success of deep neural networks in solving many fundamental computer vision tasks has largely dictated the research in the past years, but recent interest in prediction under incomplete inputs has motivated combining the extreme flexibility and expressive power of current computer vision models with structured constraints encoded by GP priors. Application areas include uncertainty quantification \citep[see discussion in][]{Blundell+Cornebise+Kavukcuoglu+Wierstra:2015,Kendall+Gal:2017}, auxiliary data fusion, and prediction under scarce data. These are instrumental for delivering practical methods and robustifying inference.

\begin{figure}[!t]
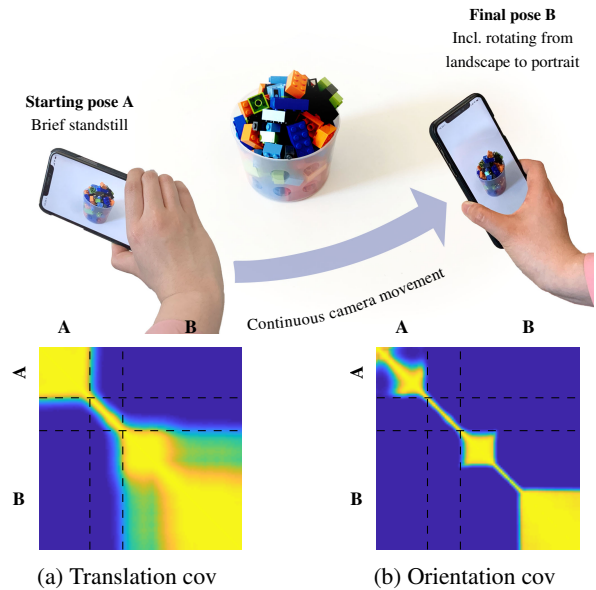

  \centering\scriptsize
  \pgfplotsset{yticklabel style={rotate=90}, ylabel style={yshift=-15pt},clip=true,scale only axis,axis on top,clip marker paths,legend style={row sep=0pt},legend columns=-1,xlabel near ticks}
  \setlength{\figurewidth}{.4\textwidth}
  \setlength{\figureheight}{.6\figurewidth}
  \begin{subfigure}{\columnwidth}
    \centering
    \resizebox{\textwidth}{!}{%
    \begin{tikzpicture}[inner sep=0]      
      \node [] at (0,0) {\includegraphics[width=\textwidth,trim=0 0 0 7cm,clip]{fig/teaser}};

      \draw [draw=none,postaction={decorate,decoration={raise=-2.5ex,text along path,text align=center,text={||Continuous camera movement}}}] (-1.5,-2) .. controls (0.1,-1.95) .. (2.35,-1);
      \node[] at (-3.5,1) {\begin{minipage}{3cm}\centering {\bf Starting pose A} \\ Brief standstill \end{minipage}};
      \node[] at (3,2.1) {\begin{minipage}{3cm}\centering {\bf Final pose B} \\ Incl.\ rotating from \\ landscape to portrait \end{minipage}};

    \end{tikzpicture}}

    \label{fig:teaser-a}
  \end{subfigure}\\[-6pt]
  \begin{minipage}{\columnwidth}
  \centering
  \begin{subfigure}{.4\textwidth}
    \centering
    \begin{tikzpicture}[inner sep=0]
       \node[] at (0,0) {\includegraphics[width=.8\textwidth]{fig/cov_trans}};       
       \node[] at (-.30\textwidth,0.48\textwidth) {\bf A};
       \node[] at (.2\textwidth,0.48\textwidth) {\bf B};       
       \node[rotate=90] at (-0.48\textwidth,.30\textwidth) {\bf A};
       \node[] at (-0.48\textwidth,-.2\textwidth) {\bf B};
       \draw[black,dashed] (-.2\textwidth,-.4\textwidth) -- (-.2\textwidth,.4\textwidth);
       \draw[black,dashed] (-.07\textwidth,-.4\textwidth) -- (-.07\textwidth,.4\textwidth);
       \draw[black,dashed] (-.4\textwidth,.2\textwidth) -- (.4\textwidth,.2\textwidth);
       \draw[black,dashed] (-.4\textwidth,.07\textwidth) -- (.4\textwidth,.07\textwidth);       
    \end{tikzpicture}   
    \caption{Translation cov}
    \label{fig:teaser-b}    
  \end{subfigure}
  \hspace*{1cm}
  \begin{subfigure}{.4\textwidth}
    \centering
    \begin{tikzpicture}[inner sep=0]
       \node[] at (0,0) {\includegraphics[width=.8\textwidth]{fig/cov_rot}};
       \node[] at (-.30\textwidth,0.48\textwidth) {\bf A};
       \node[] at (.2\textwidth,0.48\textwidth) {\bf B};       
       \node[rotate=90] at (-0.48\textwidth,.30\textwidth) {\bf A};
       \node[] at (-0.48\textwidth,-.2\textwidth) {\bf B};
       \draw[black,dashed] (-.2\textwidth,-.4\textwidth) -- (-.2\textwidth,.4\textwidth);
       \draw[black,dashed] (-.07\textwidth,-.4\textwidth) -- (-.07\textwidth,.4\textwidth);
       \draw[black,dashed] (-.4\textwidth,.2\textwidth) -- (.4\textwidth,.2\textwidth);
       \draw[black,dashed] (-.4\textwidth,.07\textwidth) -- (.4\textwidth,.07\textwidth);       
    \end{tikzpicture}    
    \caption{Orientation cov}
    \label{fig:teaser-c}
  \end{subfigure}  
  \end{minipage}
  \caption{We propose a Gaussian process prior for encoding known six degrees-of-freedom camera poses into probabilistic models. In region A, the phone starts from standstill with minor rotation at the left (high overall covariance in (a)). Between A and B, it moves to the right while rotating (low overall covariance in (a) and (b)). In B, the phone firstly stands still (high overall covariance), then rotates from portrait to landscape with minor translation (low covariance in (b), higher in (a)), and finally stands still again.}
  \label{fig:teaser}
 \vspace*{-1em}
\end{figure}

In this paper, we aim to fill a tangible gap between the theory and practice of applying probabilistic methods to certain computer vision tasks. We propose a tailored Gaussian process prior for encoding knowledge of camera poses into probabilistic models. In GPs, prior assumptions are encoded by a covariance function. As illustrated in \cref{fig:teaser}, we aim to encode the notion of {\em similarity} of camera views given the known camera movement.

In practice, the camera movement estimation is typically fused with motion information from inertial sensors. New consumer hardware in smartphones and cars typically have these capabilities built-in---Apple iPhones/iPads run ARKit and Android devices Google ARCore, both exposing real-time six degrees-of-freedom camera pose data. This readily available motion information could be utilized as priors for improving standard visual regression and classification tasks. However, typical computer vision methods operating on a stream of images consider the frames independently and merely post-process the outputs by, \eg, linear interpolation or temporal low-pass filtering. 

This paper is {\em bridging}: We emphasize the principled link between computer vision and non-parametric inference for encoding probabilistic information between camera poses, advocating for the use of more principled strategies for inter-frame reasoning in computer vision. Our contributions in this paper are:
{\em (i)}~We propose a novel view covariance function for encoding 3D camera orientation which extends the theory of GP models towards vision applications.
{\em (ii)}~We push the boundaries of GP applications in computer vision. For the first time, we use a GP model on an autoencoder to predict learnt shapes in arbitrary angles.
{\em (iii)}~We also introduce an approach to non-linear latent space interpolation in generative image models, using our view kernel.

Code and material related to this paper is available at \url{https://aaltoml.github.io/view-aware-inference}.

\section{Background}
\label{sec:background}
Gaussian processes (GPs) provide a probabilistic plug-and-play framework for specifying prior knowledge inside models. As a general-purpose machine learning paradigm they are instrumental in applications for discovering structure in signals \cite{Duvenaud:2014}, regression tasks \cite{Bui:2016}, data-efficient reinforcement learning \cite{Deisenroth:2011}, and probabilistic numerics \cite{Hennig:2015}. In theory, their applicability is only limited by the availability of prior knowlege that can be encoded.

We focus on GP models that admit the form of a {\em Gaussian process prior}
  $f(\vx) \sim \mathrm{GP}(\mu(\vx), \kappa(\vx,\vx'))$ and {\em likelihood} $\vy \mid \vf\sim\prod_{i=1}^{n} p(y_{i} \mid f(\vx_{i}))$,
where the data $\mathcal{D} = \{(\vx_i, y_i)\}_{i=1}^n$ are input--output pairs, $\mu(\vx)$ the mean, and $\kappa(\vx,\vx')$ the covariance function of the GP prior. This family covers many standard modelling problems, including regression and classification tasks.

GPs are typically associated with two issues hindering their wider use: {(\em i)}~prohibitive cubic scaling in the number of training samples $n$ and {(\em ii)}~the need for approximative inference when dealing with non-Gaussian likelihoods. Recent research has delivered methods to overcome these limitations by methods such as basis function projection \cite{Lazaro-Gredilla+Quinonero-Candela+Rasmussen+Figueiras-Vidal:2010,Hensman+Durrande+Solin:2018}, matrix structure exploiting \cite{Wilson+Nickisch:2015-ICML,Wang+Pleiss+Gardner+Tyree+Weinberger+Wilson:2019}, stochastic inference \cite{Hensman+Fusi+Lawrence:2013,Krauth+Bonilla+Cutajar+Filippone:2017}, and temporal models \cite{Sarkka+Solin+Hartikainen:2013,Solin+Hensman+Turner:2018}. The availability of GPU-accelerated software libraries such as GPflow \cite{Matthews:2017} and GPyTorch \cite{Gardner+Pleiss+Weinberger+Bindel+Wilson:2018} have recently made GP models more applicable as building blocks for larger models. Therefore, the traditional limitations are now less severe, allowing GPs to provide exciting opportunities for computer vision applications.

In this paper, the main contributions relate to the GP prior, where the {\it a~priori} assumptions are encoded by the covariance function (kernel) $\kappa(\cdot,\cdot)$. Without loss of generality, we constrain our interest to models with $\mu(\vx)=0$. Some SLAM methods exploit GP priors in $\mathrm{SE}(3)$ for continuous trajectory estimation  \cite{7353368}. For computer vision and graphics applications, recent work in kernel design has focused more on encoding the ignorance rather than the knowledge about orientation. Invariant kernels \citep[see, \eg,][]{Haasdonk+Burkhardt:2007} can robustify deep convolutional models against rotation, while translation insensitive kernels \cite{Dutordoir+Wilk+Artemev+Tomczak+Hensman:2019} can account for problems with patch similarity across images. We, however, aim to encode explicit prior knowledge about inter-image camera poses---view similarity---by crafting a view kernel that accounts for camera translation and orientation. \citet{song2009hilbert} proposed an inner product kernel between rotations, which can be regarded as a linear model in the Hilbert space, while we span a multi-dimensional periodic model in that space. This line of research also connects to distance measures between rigid bodies \cite{Mazzotti+Sancisi+Parenti-Castelli:2016}.  

Perhaps due to the two limitations mentioned earlier, GPs have not been extensively used in computer vision applications. Sufficient and necessary conditions for Gaussian kernels on metric spaces are derived in \citet{Jayasumana:2013}, with the focus on theoretical ground-work. GP priors for rigid motions applied to object tracking is extensively studied in \citet{Lang+Hirche:2017,Lang+Kleinsteuber+Hirche:2018}, which we also compare against.  There has also been previous work in combining variational autoencoders with GP priors in vision \cite{Eleftheriadis+Rudovic+Deisenroth+Pantic:2016,Casale+Dalca+Saglietti+Listgarten+Fusi:2018} and GP based latent variable models for multi-view and view-invariant facial expression recognition \cite{Eleftheriadis+Rudovic+Pantic:2015a,Eleftheriadis+Rudovic+Pantic:2015b}. In \citet{Casale+Dalca+Saglietti+Listgarten+Fusi:2018}, GPs are applied to face image modelling, where the GP accounts for the pose, and in \citet{Urtasun:2006} used them for 3D people tracking.

From an application point of view, leveraging information from consecutive views lies at the heart of many subfields in computer vision. Video analysis, multi-view methods, optical flow, visual tracking, and motion estimation and correction all directly build on the object or camera movement cues in consecutive image frames. View priors can also help in semantic processing of video \cite{Everingham+Sivic+Zisserman:2006} or depth estimation \cite{Hou+Kannala+Solin:2019,Hou+Janjua+Kannala+Solin:2021}. However, in many `one-shot' applications in visual regression and classification, the frames of the image sequence are treated as independent of one another, and typically processed with linear interpolation or low-pass filtering.

\section{Camera Pose Priors}
\label{sec:methods}
In geometric computer vision \citep[\eg,][]{Hartley+Zisserman:2003}, the standard description of a camera projection model is characterized by {\em extrinsic} and {\em intrinsic} camera parameters. The extrinsic parameters denote the coordinate system transformations from world coordinates to camera coordinates, while the intrinsic parameters map the camera coordinates to image coordinates. In the standard {\em pinhole camera} model, this corresponds to
\begin{equation}\label{eq:camera}
  \begin{pmatrix} u & v & 1 \end{pmatrix}\T \propto \MK \begin{pmatrix} \MR\T & -\MR\T \vp \end{pmatrix} \begin{pmatrix} x & y & z & 1 \end{pmatrix}\T,
\end{equation}
where $(u,v)$ are the image (pixel) coordinates, $(x,y,z) \in \R^3$ are the world coordinates, $\MK$ is the intrinsic matrix and the $\vp \in \R^3$ and  $\MR$ describe the position of the camera centre and the orientation in world coordinates respectively. From \cref{eq:camera}, given a set of fixed world coordinates and a known motion between frames the driver for changes in pixel values $(u,v)$ is the camera pose $P=\{\vp,\MR\}$.

\subsection{Kernels in $\mathrm{SE}(3)$}
In the mathematical sense, the three-dimensional camera poses belong to the special Euclidean group, $\mathrm{SE}(3)$, whose elements are called rigid motions or Euclidean motions. They comprise arbitrary combinations of translations and rotations, but not reflections. This group contains transformations represented as a translation followed by a rotation: $\mathrm{SO}(3) {\times} \mathrm{T}(3)$, where the former denotes the special orthogonal rotation group and the latter the group of translations. A camera pose $P = \{\vp,\MR\}$ is an element of this group.
We consider the orientation and translation contributions entering the prior separately: $\kappa_\mathrm{pose}(P,P') = \kappa_\mathrm{trans.}(\vp,\vp')\,\kappa_\mathrm{view}(\MR,\MR')$, since in the general case separability imposes a less informative prior. As the translation vectors reside in $\R^3$, we may directly write the translation kernel as any suitable covariance function \citep[see, \eg,][]{Rasmussen+Williams:2006,Duvenaud:2014}. An apparent first choice is the so-called squared exponential (RBF, exponentiated quadratic) covariance function:
\begin{equation}\label{eq:trans}
  \kappa(\vp,\vp') 
  = \sigma^2 \exp\left(-\frac{\|\vp-\vp'\|^2}{2\ell^2}\right),
\end{equation}
where $\sigma^2$ denotes a magnitude and $\ell>0$ is a characteristic lengthscale hyperparameter. This particular choice of covariance function encodes continuity, smoothness, and translation invariance in $\vp$. An example realization of the translation covariance matrix 
is visualized in \cref{fig:teaser-b}.

\begin{figure*}[!t]
  \centering\scriptsize
  \pgfplotsset{yticklabel style={rotate=90}, ylabel style={yshift=-15pt},clip=true,scale only axis,axis on top,clip marker paths,legend style={row sep=0pt},legend columns=-1,xlabel near ticks,clip=false}
  \setlength{\figurewidth}{.2\textwidth}
  \setlength{\figureheight}{\figurewidth}
  \captionsetup[subfigure]{justification=centering}
  \begin{subfigure}[b]{.28\textwidth}
    \centering
%
%
\begin{tikzpicture}

\begin{axis}[%
point meta min=0,
point meta max=3.14159265358979,
axis on top,
xmin=-6.29235784047474,
xmax=6.29235784047474,
xtick={-6.28318530717959,-3.14159265358979,0,3.14159265358979,6.28318530717959},
xticklabels={{$-2\pi$},{$-\pi$},{0},{$\pi$},{$2\pi$}},
xlabel={$\theta_1$},
y dir=reverse,
ymin=-6.29234446943495,
ymax=6.29234446943495,
ytick={-6.28318530717959,-3.14159265358979,0,3.14159265358979,6.28318530717959},
yticklabels={{$-2\pi$},{$-\pi$},{0},{$\pi$},{$2\pi$}},
ylabel={$\theta_2$},
axis background/.style={fill=white},
legend style={legend cell align=left,align=left,draw=white!15!black},
width=\figurewidth,
height=\figureheight
]
\addplot [forget plot] graphics [xmin=-6.29235784047474,xmax=6.29235784047474,ymin=-6.29234446943495,ymax=6.29234446943495] {./fig/geodesic-1.png};
\end{axis}
\end{tikzpicture}%
    \vspace*{-1em}
    \caption{Geodesic \ref{addplot:distance0}}
    \label{fig:distance-a}
  \end{subfigure}
  \hspace*{\fill}
  \begin{subfigure}[b]{.28\textwidth}
    \centering
%
%
\begin{tikzpicture}

\begin{axis}[%
point meta min=0,
point meta max=3.14159265358979,
axis on top,
xmin=-6.29235784047474,
xmax=6.29235784047474,
xtick={-6.28318530717959,-3.14159265358979,0,3.14159265358979,6.28318530717959},
xticklabels={{$-2\pi$},{$-\pi$},{0},{$\pi$},{$2\pi$}},
xlabel={$\theta_1$},
y dir=reverse,
ymin=-6.29234446943495,
ymax=6.29234446943495,
ytick={-6.28318530717959,-3.14159265358979,0,3.14159265358979,6.28318530717959},
yticklabels={{$-2\pi$},{$-\pi$},{0},{$\pi$},{$2\pi$}},
ylabel={$\theta_2$},
axis background/.style={fill=white},
legend style={legend cell align=left,align=left,draw=white!15!black},
width=\figurewidth,
height=\figureheight
]
\addplot [forget plot] graphics [xmin=-6.29235784047474,xmax=6.29235784047474,ymin=-6.29234446943495,ymax=6.29234446943495] {./fig/quaternion-1.png};
\end{axis}
\end{tikzpicture}%
    \vspace*{-1em}
    \caption{Quaternion \ref{addplot:distance2}}
    \label{fig:distance-b}    
  \end{subfigure}  
  \hspace*{\fill}
  \begin{subfigure}[b]{.4\textwidth}
    \setlength{\figureheight}{\figurewidth}  
    \setlength{\figurewidth}{.8\textwidth}  
    \pgfplotsset{y axis line style={draw opacity=0}}
    \centering
    \input{./fig/diag-dist.tex}
    \vspace*{-1em}
    \caption{Distance (diagonal, $\theta_1 = \theta_2$)}
    \label{fig:distance-c}    
  \end{subfigure}\\[-6pt]
  \begin{subfigure}[b]{.28\textwidth}
    \centering
%
%
\begin{tikzpicture}

\begin{axis}[%
point meta min=0,
point meta max=3.14159265358979,
axis on top,
xmin=-6.29235784047474,
xmax=6.29235784047474,
xtick={-6.28318530717959,-3.14159265358979,0,3.14159265358979,6.28318530717959},
xticklabels={{$-2\pi$},{$-\pi$},{0},{$\pi$},{$2\pi$}},
xlabel={$\theta_1$},
y dir=reverse,
ymin=-6.29234446943495,
ymax=6.29234446943495,
ytick={-6.28318530717959,-3.14159265358979,0,3.14159265358979,6.28318530717959},
yticklabels={{$-2\pi$},{$-\pi$},{0},{$\pi$},{$2\pi$}},
ylabel={$\theta_2$},
axis background/.style={fill=white},
legend style={legend cell align=left,align=left,draw=white!15!black},
width=\figurewidth,
height=\figureheight
]
\addplot [forget plot] graphics [xmin=-6.29235784047474,xmax=6.29235784047474,ymin=-6.29234446943495,ymax=6.29234446943495] {./fig/separable-1.png};
\end{axis}
\end{tikzpicture}%
    \vspace*{-1em}
    \caption{Separable \ref{addplot:distance1} \\ (Euler angles)}
    \label{fig:distance-d}    
  \end{subfigure}
  \hspace*{\fill}
  \begin{subfigure}[b]{.28\textwidth}
    \centering
%
%
\begin{tikzpicture}

\begin{axis}[%
point meta min=0,
point meta max=3.14159265358979,
axis on top,
xmin=-6.29235784047474,
xmax=6.29235784047474,
xtick={-6.28318530717959,-3.14159265358979,0,3.14159265358979,6.28318530717959},
xticklabels={{$-2\pi$},{$-\pi$},{0},{$\pi$},{$2\pi$}},
xlabel={$\theta_1$},
y dir=reverse,
ymin=-6.29234446943495,
ymax=6.29234446943495,
ytick={-6.28318530717959,-3.14159265358979,0,3.14159265358979,6.28318530717959},
yticklabels={{$-2\pi$},{$-\pi$},{0},{$\pi$},{$2\pi$}},
ylabel={$\theta_2$},
axis background/.style={fill=white},
legend style={legend cell align=left,align=left,draw=white!15!black},
width=\figurewidth,
height=\figureheight
]
\addplot [forget plot] graphics [xmin=-6.29235784047474,xmax=6.29235784047474,ymin=-6.29234446943495,ymax=6.29234446943495] {./fig/trace-1.png};
\end{axis}
\end{tikzpicture}%
    \vspace*{-1em}
    \caption{Non-separable \ref{addplot:distance3} \\ (with rotation matrix)}
    \label{fig:distance-e}    
  \end{subfigure}    
  \hspace*{\fill}
  \begin{subfigure}[b]{.4\textwidth}
    \setlength{\figureheight}{\figurewidth}  
    \setlength{\figurewidth}{.8\textwidth}
    \pgfplotsset{y axis line style={draw opacity=0}}    
    \centering
    \input{./fig/distance.tex}
    \vspace*{-1em}
    \caption{Distance \\ (when $\theta_2 \equiv 0$)}
    \label{fig:distance-f}    
  \end{subfigure}
  \vspace*{-6pt}
  \caption{Characterization of differences between different orientation distance measures.     {\bf (Left)}~Distance matrices between two degrees-of-freedom rotations (for simpler visualization) with scale $0$~\protect\includegraphics[width=1cm]{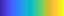}~$\pi$. (a)~shows the geodesic distance (see \cref{eq:d-rot}), (b)~the quaternion norm distance (see \cref{eq:d-quat-norm}), (d)~the separable periodic distance (from \cref{eq:per-sep}), and (e)~the non-separable orientation distance (from \cref{eq:view}). {\bf (Right)}~Distance evaluations along the diagonal and when $\theta_2 \equiv 0$, showing that in 1D (d) and (e) coincide, while (e) is symmetric in 2D/3D.}
  \label{fig:distances}
  \vspace*{-1em}    
\end{figure*}

\subsection{View Orientation Kernels}
\label{sec:view-kernels}
Since translations can be considered directly, our main interest is formulating a proper orientation covariance function in $\mathrm{SO}(3)$. Here, the first choice could be to leverage the standard periodic kernel, which can be derived following \citet{MacKay:1998}: Given a valid covariance function $\kappa(\vu,\vu')$, we can introduce a non-linear mapping $\vx \mapsto \vu(\vx)$, through which to define a new covariance function $\kappa'(\vx, \vx') \triangleq \kappa(\vu(\vx), \vu(\vx'))$. The standard periodic kernel \citep[\cf,][]{Rasmussen+Williams:2006} is usually derived by the mapping $\theta \mapsto \vu$ that warps $\theta$ to the unit circle: $\vu(\theta) = (\cos(\theta), \sin(\theta))$. Combining this with the covariance function in \cref{eq:trans} gives
\begin{equation}\label{eq:per}
  \kappa(\theta,\theta') = \exp\left(-\frac{2\sin^2((\theta-\theta')/2)}{\ell^2} \right),
\end{equation}
which can be used for imposing a periodic prior over inputs $\theta \in \R$. We aim to extend this 1D standard periodic kernel to 3D rotations \citep[see also][]{hamsici2008rotation}.

\paragraph{Euler angle formalism}  Assuming Euler angles $\vtheta = (\theta_1, \theta_2, \theta_3)$  to be fully separable, we can extend \cref{eq:per} to 3D rotations directly. This would correspond to a separable view kernel (see \cref{fig:distance-d} for the corresponding distance function):
\begin{equation}\label{eq:per-sep}
  \kappa(\vtheta,\vtheta') 
  = \prod_{j=\{1,2,3\}} \exp\bigg(-\frac{2\sin^2((\theta_j-\theta_j')/2)}{\ell_j^2} \bigg).
\end{equation}
This, however, can suffer from issues related to Euler angles like possibly singular representations and gimbal lock (loss of one degree of freedom, see, \eg, \cite{Diebel:2006,Featherstone:2014}), and should thus be avoided as an internal representation of orientation.

\paragraph{Quaternion formalism} 
Instead of Euler angles, common representations for orientation are given in terms of rotation matrices or quaternions. The set of unit quaternions, $\vq = (q_\mathrm{w},q_\mathrm{x},q_\mathrm{y},q_\mathrm{z})$, s.t.~$\|\vq\| \equiv 1$, forms the 3D rotation group $\mathrm{SO}(3)$ covering the $\mathrm{S}^3$ sphere.
In order to seek a similar, but higher-dimensional, form of \cref{eq:per}, the quaternion representation can directly be used as a mapping. This would make sense, as the derivation of the standard periodic covariance function can be viewed as a mapping onto the complex plane and quaternions represent a 4D extension of complex numbers. So we may define the distance between quaternions $\vq_1$ and $\vq_2$ as the norm of their difference:
\begin{equation}\label{eq:d-quat-norm}
  d_\mathrm{quat}(\vq_1, \vq_2) = 2\|\vq_1 - \vq_2\|.
\end{equation}
The quaternion model has previously been discussed by \citet{Lang+Hirche:2017} and \citet{Lang+Kleinsteuber+Hirche:2018}. However, the resulting covariance function is not well-behaved in all orientations---due to non-uniqueness of quaternions---as can be seen from \cref{fig:distance-b} (or \cref{fig:covariances} in the Appendix), where full-turn ($2\pi$) correlations are close to zero.

\paragraph{Rotation matrix formalism}
The peculiarities with the previous formulations, as visualized in \cref{fig:distances} (and \cref{fig:covariances} in the Appendix), acted as a motivation to seek a more principled generalization of the periodic covariance function with rotation matrices.   Since there is no direct way to use a rotation matrix as a mapping to extend  \cref{eq:per}, we consider the geodesic (arc) distance.
Considering the eigendecomposition of $\MR$ that define the rotation axis and angle (see \cref{app:geodesic}),
we have the geodesic distance defined by rotation matrices $\MR$:
\begin{equation}\label{eq:d-rot}
d_\mathrm{g}(\MR, \MR') = \mathrm{arccos}\bigg(\frac{1}{2}(\mathrm{tr}(\MR\T \MR')-1)\bigg).
\end{equation}
To derive the  3D counterpart of the standard periodic kernel, a Taylor expansion (see \cref{app:deriv}) for the geodesic distance around the origin gives a mapping $d_\mathrm{g}(\MR, \MR')\approx\sqrt{\mathrm{tr}(\MI - \MR\T\MR')}$ (visualized in \cref{fig:distance-e}) that we use for the non-separable covariance function:
\begin{equation}\label{eq:view}
  \kappa_\mathrm{view}(\MR,\MR') = \exp\left(-\frac{\mathrm{tr}(\MI - \MR\T\MR')}{2\ell^2} \right).
\end{equation}
This proposed 3D kernel \cref{eq:view} gives the standard periodic kernel as a special case where there is only rotation around one of the axes (see \cref{fig:distance-f} and \cref{app:link}). Moreover, the proposed \cref{eq:view} may be generalized to $\kappa_\mathrm{view}(\MR,\MR') = \exp(-\frac{1}{2}\,\mathrm{tr}(\MLambda - \MR\T\MLambda\MR'))$, where $\MLambda = \mathrm{diag}(\ell_\mathrm{x}^{-2},\ell_\mathrm{y}^{-2},\ell_\mathrm{z}^{-2})$, which can account for different characteristic scaling per axis flexibly.  (NB: The $\ell$s are coupled and its interpretation is not as straightforward as scaling for the respective axes)

To summarize, we propose the non-separable orientation covariance function $\kappa_\mathrm{view}(\cdot,\cdot)$ that preserves a symmetric correlation structure around origin (like the geodesic model), does not suffer from the degeneracy of Euler angles, and generalizes the gold-standard (one-dimensional) periodic kernel to high-dimensional rotations.

\section{Application Experiments}
\label{sec:experiments}
In the experiments, we show examples of real-world applications of the view kernel in probabilistic view synthesis. In the first experiment, we extend the GP variational autoencoder model with our view kernel for a view synthesis task. The second experiment is concerned with latent space interpolation for human face modelling, showcasing the general applicability of the kernel. Further examples and comparisons are included in the appendix (see \cref{sec:track}).

\begin{figure*}[!t]
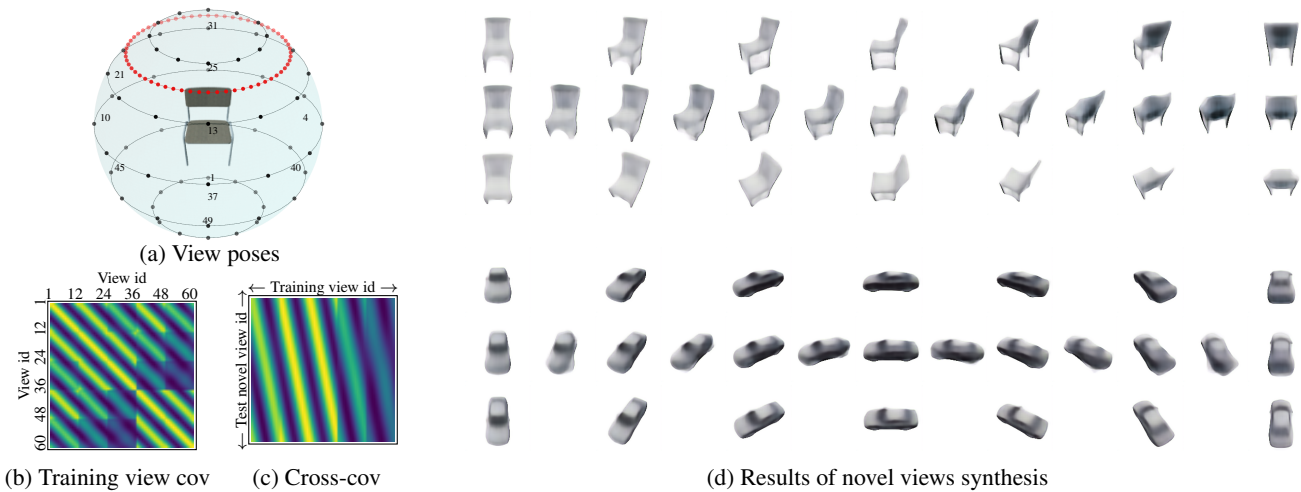

  \centering\scriptsize
  \pgfplotsset{yticklabel style={rotate=90}, ylabel style={yshift=-15pt},clip=true,scale only axis,axis on top,clip marker paths,legend style={row sep=0pt},legend columns=-1,xlabel near ticks,clip=false}
  \setlength{\figurewidth}{.2\textwidth}
  \setlength{\figureheight}{\figurewidth}
 \begin{minipage}[b]{.34\textwidth}
  \begin{subfigure}[b]{.9\textwidth} 
    \centering 
    \scalebox{.65}{%
    \begin{tikzpicture}[inner sep=0]
      \node[] at (0,0) {\includegraphics[width=.9\textwidth]{./fig/chairs/chair60}};
      \node at (.1,-1.1) {\tiny 1};
      \node at (2.0,.12) {\tiny 4};
      \node at (-2.1,.12) {\tiny 10};
      \node at (.1, -.15) {\tiny 13};      
     
      \node at (-1.8,1.0) {\tiny 21};
     \node at (.1,1.15 ) {\tiny 25};
      \node at (.1,2 ) {\tiny 31};
        
      \node at (.1,-1.5) {\tiny 37};
     \node at (1.8,-.9) {\tiny 40};
       \node at (-1.8,-.9) {\tiny 45};

\node at (0,-2) {\tiny 49};    
    \end{tikzpicture}}
    \vspace*{-1em}
    \caption{View poses}  
    \label{fig:chairs-a}    
  \end{subfigure}\\
  \begin{subfigure}[b]{.45\textwidth} 
    \centering\tiny
    \begin{tikzpicture}[inner sep=0]
       \foreach \i in {1,12,24,36, 48, 60} {
         \node[] at ({-0.35*\textwidth+\i/60*0.7*\textwidth-0.7*\textwidth/36},0.4\textwidth) {\tiny \i};
         \node[rotate=90] at (-0.4\textwidth,{0.35*\textwidth-\i/60*0.7*\textwidth+0.7*\textwidth/36}) {\tiny \i};
       }
       \node[] at (0,0.48\textwidth) {View id};
       \node[rotate=90] at (-0.48\textwidth,0) {View id};       
       \node[draw=black,inner sep=1pt] at (0,0) {\includegraphics[width=.7\textwidth]{fig/chairs/view_cov}};
    \end{tikzpicture}

    \caption{Training view cov}  
    \label{fig:chairs-b}    
  \end{subfigure}
  \begin{subfigure}[b]{.45\textwidth} 
    \centering\tiny

    \begin{tikzpicture}[inner sep=0]
       \node[] at (0,0.4\textwidth) {$\leftarrow$ Training view id $\rightarrow$};
       \node[rotate=90] at (-0.4\textwidth,0) {$\leftarrow$ Test novel view id $\rightarrow$};
       \node[draw=black,inner sep=1pt] at (0,0) {\includegraphics[width=.7\textwidth]{fig/chairs/view_cov_test}};
    \end{tikzpicture}

    \caption{Cross-cov}  
    \label{fig:chairs-c}
  \end{subfigure}
  \end{minipage}
  \begin{minipage}[b]{.65\textwidth}
  \begin{subfigure}[b]{.95\textwidth}
    \tiny
    \centering
    \setlength{\figurewidth}{.079\textwidth}
    \setlength{\figureheight}{.95\figurewidth}    
    
    \begin{tikzpicture}[inner sep=0]
      \foreach \x [count=\i] in {0012,0013,0014, 0015,0016,0017,0018} 
      {

        \node[draw=white,fill=black!20,minimum size=\figurewidth,inner sep=0pt]
        (\i) at ({2*\figurewidth*\i},{-\figureheight*1})
        {\includegraphics[width=\figurewidth]{./fig/chairs/oos/reco/\x}};
      }
      \end{tikzpicture}
      
    \begin{tikzpicture}[inner sep=0]
      \foreach \x [count=\i] in {0000,0002,0005, 0007,0010,0012,0015,0017,0020,
0022,0025,0027,0030} 
      {

        \node[draw=white,fill=black!20,minimum size=\figurewidth,inner sep=0pt]
        (\i) at ({\figurewidth*\i},{-\figureheight*1})
        {\includegraphics[width=\figurewidth]{./fig/chairs/oos/pred/\x}};
      }
    \end{tikzpicture}

    \begin{tikzpicture}[inner sep=0]
      \foreach \x [count=\i] in {0024,0025,0026, 0027,0028,0029, 0030} 
      {

        \node[draw=white,fill=black!20,minimum size=\figurewidth,inner sep=0pt]
        (\i) at ({2*\figurewidth*\i},{-\figureheight*1})
        {\includegraphics[width=\figurewidth]{./fig/chairs/oos/reco/\x}};
      }
    \end{tikzpicture}
    \vspace*{0.7em} 
    
    \begin{tikzpicture}[inner sep=0]
      \foreach \x [count=\i] in {0012,0013,0014, 0015,0016,0017,0018} 
      {

        \node[draw=white,fill=black!20,minimum size=\figurewidth,inner sep=0pt]
        (\i) at ({2*\figurewidth*\i},{-\figureheight*1})
        {\includegraphics[width=\figurewidth]{./fig/chairs/oos_car/reco/\x}};
      }
      \end{tikzpicture}
    \begin{tikzpicture}[inner sep=0]
      \foreach \x [count=\i] in {0000,0002,0005, 0007,0010,0012,0015,0017,0020,
0022,0025,0027,0030} 
      {

        \node[draw=white,fill=black!20,minimum size=\figurewidth,inner sep=0pt]
        (\i) at ({\figurewidth*\i},{-\figureheight*1})
        {\includegraphics[width=\figurewidth]{./fig/chairs/oos_car/pred/\x}};
      }
    \end{tikzpicture}

    \begin{tikzpicture}[inner sep=0]
      \foreach \x [count=\i] in {0024,0025,0026, 0027,0028,0029, 0030} 
      {

        \node[draw=white,fill=black!20,minimum size=\figurewidth,inner sep=0pt]
        (\i) at ({2*\figurewidth*\i},{-\figureheight*1})
        {\includegraphics[width=\figurewidth]{./fig/chairs/oos_car/reco/\x}};
      }
    \end{tikzpicture}
    
    \caption{Results of novel views synthesis}
    \label{fig:chairs-novel}    
  \end{subfigure}

\end{minipage}

  \caption{ShapeNet experiments with a GPPVAE. (a)~Visualization of the 60 view angles (black dots) in the training data. (b)~The cov matrix for 60 training views. (c)~The cross-cov matrix for the test novel views (red dots in (a)) and training views (black dots). (d)~We experimented with both chairs and cars. Our proposed kernel allows to predict arbitrary views which are not presented in training data. For each category, the first (elevation $30^{\circ}$) and the third row (elevation $60^{\circ}$) show predictions for angles found in the training set, while the whole second row shows predictions for angles {\bf not} found in the training set (red dots in (a)).}
  \label{fig:chairs}

\end{figure*}

\subsection{View Synthesis with a GP Prior VAE}
\label{sec:gppvae}
We consider the task of using a variational autoencoder (VAE) to predict how objects look in orientations that are not in the training set. We first describe how the problem was previously addressed by \citet{Casale+Dalca+Saglietti+Listgarten+Fusi:2018} with the Gaussian Process Prior Variational Autoencoder (GPPVAE), explain a major limitation in this approach, and then overcome this limitation with our kernel.
GPPVAE is a {\em fully probabilistic} model that captures correlations in both object identities and views by leveraging covariance structure in latent space. The kernel defines a prior for latent code $\vz$. Given an object ID and view angle, the encoder and GP posterior predict the posterior $\vz$. Intuitively, the prediction is based on the relation between training samples.

Given training images $\mathcal{Y}$, training object feature vectors $\mathcal{X}$, and training views $\mathcal{P}$, the predictive posterior for an image $\vy_\star$ for an object with features $\vx_\star$ seen from a view $P_\star$ is given \citep[see detailed presentation in][]{Casale+Dalca+Saglietti+Listgarten+Fusi:2018} by 
\begin{align}
 & p(\vy_\star \mid \vx_\star, \mathcal{Y},\mathcal{X},\mathcal{P}) {\approx} \nonumber \\
& \hspace*{-3pt} \int \hspace*{-3pt}\underbrace{p(\vy_\star \mid \vz_\star)}_\text{\tiny decode prediction} \hspace*{-2pt} 
  \underbrace{p(\vz_\star \mid \vx_\star, P_\star, \mathcal{Z}, \mathcal{X}, \mathcal{P})}_\text{\tiny GP predictive posterior} 
  \hspace*{-8pt}\underbrace{q(\mathcal{Z} \mid \mathcal{Y})}_\text{\tiny encode training data} \hspace*{-8pt} \, 
  \mathrm{d}\vz_\star \, \mathrm{d}\mathcal{Z},\hspace*{-4pt}
\end{align}
where $\vz_\star$ are the predicted latent representations and $\mathcal{Z}$ are latent representations of training images. Given fixed views and objects, the task of GPPVAE is to predict images $\vy_\star$ for an object in the view $P_\star$ that remained unobserved.  

However, though \citet{Casale+Dalca+Saglietti+Listgarten+Fusi:2018} present the task as `out-of-sample' prediction, their approach of brute learning the covariance does not support arbitrary 3D angles. Rather, it is defined based on the assumption that all query views in the test set have already been observed for at least one object in the training set. When that assumption does not hold, only a fixed number of 3D rotations are available. In GPPVAE, all experiments only consider rotations in one dimension, modelled with the 1D standard periodic kernel or the fully-learned kernel. The 1D standard periodic kernel cannot handle 3D rotations and the fully-learned kernel can only capture the correlations within fixed training views. In contrast, our proposed kernel that extends the 1D standard periodic kernel to $\mathrm{SO}(3)$ can work with arbitrary 3D angles.

To showcase our kernel with 3D rotations, we carried out an experiment with ShapeNet~\cite{Chang+Funkhouser+Guibas+Hanrahan+Huang+Li+Savarese+Savva+Song+Su+others:2015} 3D chair models at $128{\times}128$ resolution.  We use 1660 different chairs in total. For each object, we render images from 60 fixed views, considering both azimuth angles ($0^{\circ}, 30^{\circ}, 60^{\circ}, \ldots, 330^{\circ}$) and elevation angles ($ 0^{\circ}, 30^{\circ}, 60^{\circ},  -30^{\circ},  -60^{\circ}$). The camera view angles are shown in \cref{fig:chairs-a}. We randomly selected 80\% images for training (81,312 images), 10\% for validation (10,164 images) and 10\% for testing (10,164 images). Following original GPPVAE, we compute the view covariance based only on orientation angles (cameras at fixed radius from the object centre; translation seen as function of orientation). For the object covariance, we use a linear kernel between learned object features. The resulting composite kernel $\kappa(\vx, \MR; \vx', \MR')$ expresses the covariance between two chair images in terms of the relative view orientation between orientations $\MR$ and $\MR'$ and object feature vectors $\vx$ and $\vx'$:
\begin{equation}\label{eq:viewx}
  \hspace*{-4pt}\kappa(\vx, \MR; \vx', \MR') = \underbrace{\vx\T\vx'\vphantom{\frac{1}{2}}}_{\text{object}}\underbrace{\exp\!\big(-\frac{1}{2}\mathrm{tr}(\MLambda - \MR\T\MLambda\MR')\big)}_{\text{view}},
\end{equation}
where $\MLambda = \mathrm{diag}(\ell_\mathrm{x}^{-2},\ell_\mathrm{y}^{-2},\ell_\mathrm{z}^{-2})$ and we learn the lengthscale hyperparameters $\ell_\mathrm{x}, \ell_\mathrm{y}, \ell_\mathrm{z}$ as part of the training. Due to rich variability in chair shapes, we consider a higher rank ($M=128$) than the original setup for the object covariance (see \cref{app:gppvae} for details).
We first experiment on same task as GPPVAE (in-sample evaluation). For the proposed view kernel, the MSE is $0.025{\pm}0.012$, which still has slightly better performance than the fully-learned view-covariance matrix as in \citet{Casale+Dalca+Saglietti+Listgarten+Fusi:2018} ($0.026{\pm}0.012$). This also shows that encoding the information through a view kernel (with only hyperparameters to learn), rather than through brute free-form optimization, is sensible.

\begin{figure*}[t!]
  \centering\scriptsize
  \pgfplotsset{yticklabel style={rotate=90}, ylabel style={yshift=-15pt},clip=true,scale only axis,axis on top,clip marker paths,legend style={row sep=0pt},legend columns=-1,xlabel near ticks}
  \begin{subfigure}{\textwidth}
    \setlength{\figurewidth}{.333\textwidth}
    \begin{tikzpicture}[inner sep=0]
      \foreach \x [count=\i] in {a,b,c} 
      {
        \node[inner sep=0pt]
        (\i) at ({\figurewidth*\i},{0})
        {\includegraphics[width=\figurewidth]{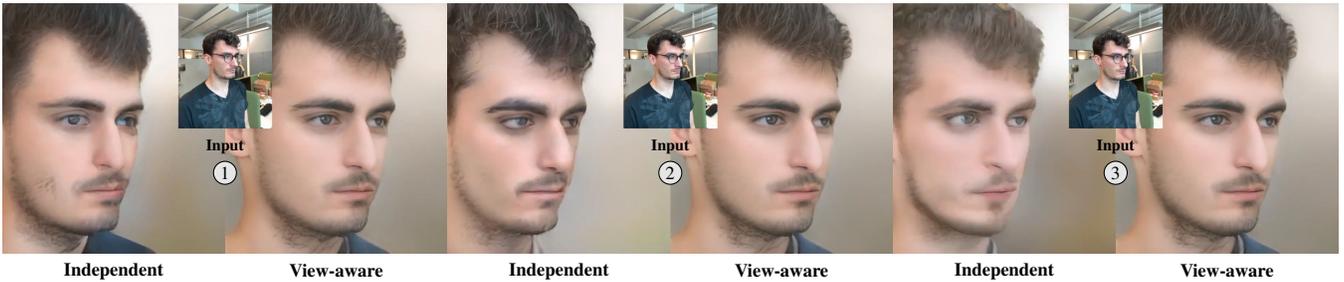}};
        
        \node[inner sep=1pt,shape=circle,draw=black,fill=black!10] at ({\figurewidth*\i},{-.1*\figurewidth}) {\i};
        \node[inner sep=0pt] at ({\figurewidth*\i},{-0.04*\figurewidth}) {\tiny \bf Input};
        \node[inner sep=0pt] at ({\figurewidth*\i-0.25*\figurewidth},{-0.32*\figurewidth}) {\scriptsize \bf Independent};  
        \node[inner sep=0pt] at ({\figurewidth*\i+0.25*\figurewidth},{-0.32*\figurewidth}) {\scriptsize \bf View-aware};        
        
      }
    \end{tikzpicture}\\[-6pt]~

  \end{subfigure}\\[3pt]
  \vspace*{-2em}
  \caption{View-aware manipulations in the latent space of StyleGAN~\cite{Karras+Laine+Aila:2019}. Example of denoising of the GAN reconstructions for consecutive frames (note the noisy independent reconstructions) in a longer video, where every frame is treated as a noisy observation. See the supplement for video examples.
  }
  \label{fig:denoising-interpolation}

\end{figure*}

\cref{fig:chairs-novel} demonstrates the capability of our kernel for novel view predictions conditioned on an object ID, with truly `out-of-sample' views (novel viewpoints in red in \cref{fig:chairs-a}). The closest views within the training set are also visualized, which demonstrates that our model has learned to disentangle view and content by the aid of the view prior. The qualitative results on ShapeNet cars also show the generalizability.

We evaluate MSEs for the novel view prediction for each kernel, using the trained lengthscale and magnitude hyperparameters from the view kernel (the parameters have the same interpretation across kernels). The practical degeneracy of the separable kernels (based on Euler angles) and quaternion kernels can make training unstable. For our non-separable view kernel we get an MSE of 0.036. Given the hyperparameters trained with the non-separable model, the separable model performs almost equally well. The quaternion distance kernel fails at this task (MSE 0.058).

\subsection{Robust Interpolation for Face Reconstruction}
\label{sec:face}
As a second example of inter-frame reasoning, we consider view-aware GP interpolation in the latent space of a Generative Adversarial Network \citep[GAN,][]{Goodfellow+Pouget-Abadie+Mirza+Xu+Warde-Farley+Ozair+Courville+Bengio:2014} for face generation. A GAN incorporates a generator network that acts as a feature extractor, allowing an image to be represented by a low-dimensional latent code.
By utilizing the pose information of the view-aware kernel, we can do GP regression in the latent space. The data comprises short video sequences of faces of four volunteers captured by an Apple iPhone~XS. We used a custom app for capturing the video stream ($1440{\times}1920$ at 60~Hz) interleaved with camera poses from the Apple ARKit API.

In absence of a built-in encoder, as in case of most GANs, we use an optimization setup to find out the best latent code for an image $j$ \citep[similarly to][]{abdal:2019}. The traditional approach has been to learn these codes from i.i.d.\ training data, and under the assumption that we essentially have only a single `observation' of each entity that the image represents. We now relax this assumption and consider the more general case where we postulate, for each input image frame, the existence of a hidden `correct' latent code $\vf_j \in \R^d$ that encodes both the time-invariant aspect (face identity) and the time-dependent aspect (pose of the face), and then re-interpret each latent code produced by an encoder or optimizer as a noisy `observation' $\vy_j \in \R^d$ of the correct code. 
Consider the case of images that depict a face with fixed identity.
We cast this as a GP regression problem in which each latent dimension, $i=1,2,\ldots,d$ is independent. The likelihood is $y_{j,i} = f_i(P_j) + \varepsilon_{j,i}$, $\varepsilon_{j,i} \sim \mathrm{N}(0,\sigma_\mathrm{n}^2)$, for frames $j=1,2,\ldots,n$. The GP prior is over the camera poses $P_j$: $f_i(P) \sim \mathrm{GP}(0,\kappa_\mathrm{view}(P,P'))$. Solving these independent GP regression problems only requires inverting one $n{\times}n$ covariance matrix, which makes inference fast.
We use two or more images of a sequence to predict the expected latent code, $\mathrm{E}[\vf(P_j) \mid \mathcal{D}]$, for any image in the sequence, without necessarily ever running that image through the encoder. We can apply these predictions in several ways, here focusing separately on noise reduction (leveraging all available image frames) and view synthesis (leveraging as few as two frames).

\begin{figure*}[!t]
  \centering\scriptsize
  \setlength{\figurewidth}{.14\textwidth}
  \setlength{\figureheight}{\figurewidth}
  \resizebox{\textwidth}{!}{%
  \begin{tikzpicture}[inner sep=0]

    \foreach \y [count=\j] in {fig/video-interpolation/face04/interpolation_comp/aligned,fig/video-interpolation/face04/interpolation_comp/stylegan,fig/video-interpolation/face04/interpolation_comp/linear,fig/video-interpolation/face04/stylegan_interpolations, fig/video-interpolation/face04/stylegan_interpolation_std} {

      \foreach \x [count=\i] in {0001,0051,0101,0151,0201,0251,0301,0356} {
          \node[draw=white,fill=black!20,minimum size=\figurewidth,inner sep=0pt]
          (\i) at ({-\figurewidth*\i},{-1*\figureheight*\j})
          {\includegraphics[width=\figurewidth]{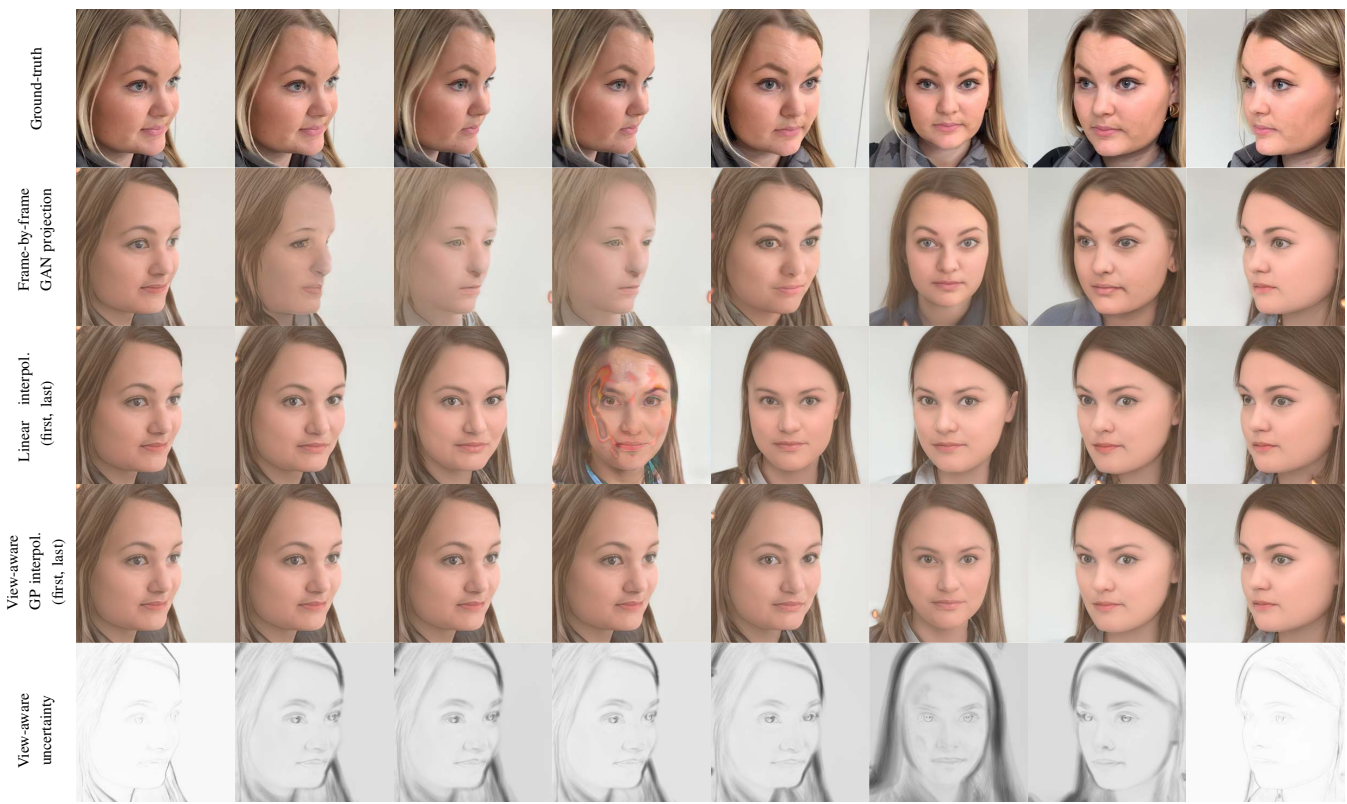}};
      }
    }

    \node[rotate=90,text width=0.8*\figureheight,align=center] at (-8.75*\figurewidth,-1*\figureheight) {\scriptsize Ground-truth};
    \node[rotate=90,text width=0.9*\figureheight,align=center] at (-8.75*\figurewidth,-1.95*\figureheight) {\scriptsize Frame-by-frame GAN projection};

    \node[rotate=90,text width=0.8*\figureheight,align=center] at (-8.75*\figurewidth,-3.05*\figureheight) {\scriptsize \mbox{~~Linear~~} \mbox{interpol.} \mbox{(first, last)}};
    \node[rotate=90,text width=0.9*\figureheight,align=center] at (-8.75*\figurewidth,-4.05*\figureheight) {\scriptsize View-aware \mbox{GP interpol.} \mbox{(first, last)}};
    \node[rotate=90,text width=0.9*\figureheight,align=center] at (-8.75*\figurewidth,-5.05*\figureheight) {\scriptsize View-aware \mbox{uncertainty}};
  \end{tikzpicture}}

  \caption{View-aware GP interpolation between two input frames: {\bf Row~\#1:}~Frames separated by equal time intervals from a camera run, aligned on the face. {\bf Row~\#2:}~Each frame independently GAN reconstructions. {\bf Row~\#3:}~Linear interpolation of the intermediate frames in GAN latent space between first and last frame (note the lost azimuth angle). {\bf Row~\#4:}~Interpolation in GAN latent space between first and last frame by our view-aware GP prior. {\bf Row~\#5:}~Per-pixel GP posterior uncertainty visualized in the form of standard deviation of the prediction at the corresponding time step. Heavier shading indicates higher uncertainty around the mean trajectory. See the supplement for video examples.}
  \label{fig:interpolation_comp}
  \vspace*{-1em}
\end{figure*}

We demonstrate this approach in the $18 {\times} 512$ latent space of StyleGAN \cite{Karras+Laine+Aila:2019} based on four image sequences, each depicting a specific face identity (see \cref{fig:interpolation_comp} and the supplement). We find the `observed' latent codes using an optimizer, leveraging VGG16 feature projections \cite{simonyan:2015,puzer:2019}. Separately for each face identity, our method infers the `correct' latent codes for each pose. The GAN generator then decodes those back to $1024 {\times} 1024$ image space. The values for the three hyperparameters were chosen to $\sigma^2 = 0.1$, $\ell = 1.098$, and $\sigma_\mathrm{n}^2 = 0.0001$ (pre-trained on an independent task w.r.t.\ marginal likelihood). Even if the GAN encoding produced stable results, the considerable slowness of finding the latent codes by optimization (in range of minutes per single image) motivates the present approach, as we now need to encode only a small subset of frames and match the camera movement by GP prediction.

\paragraph{Noise reduction} Given a sequence of images of the same object, we can use the encoder (optimizer) to find the corresponding latent codes.
As we decode the codes back to individual images, they are mutually inconsistent (no temporal consistency). The issue may not be clear when visually examining single frames, but it is plain when the frames are combined into a video (see the supplement for video examples). We `denoise' the sequence of latent codes with GP regression, and decode the new sequential images as video, making it smoother and reducing artifacts. \cref{fig:denoising-interpolation} shows three consecutive input frames from a video and their respective independent GAN reconstructions. Partly due to the tilted angle, the quality and preservation of identity in face reconstructions for independent frames varies. GP regression with our view-aware prior makes the motion smooth and preserves the identity better throughout the video. The smoothness can be measured using the mean difference of the learned perceptual image path similarity metric \citep[LPIPS,][]{zhang2018} between consequtive frames, considerably smaller for the GP interpolation using all frames (the LPIPS-$\Delta$ in \cref{tbl:lpips}).

\begin{table}[!t]
  \centering\small\scriptsize
  \caption{LPIPS similarities between ground-truth and frames generated with different methods, center-cropped, using camera runs on 4 face identities ($N=1570$). {\bf Smaller is better.} 
  \label{tbl:lpips}}
  \vspace*{-2.5em}%
  \begin{tabular*}{\columnwidth}{@{\extracolsep{\fill}} lccc}
  \toprule
  Reconstruction mode & LPIPS (mean) & \hspace*{-1em}LPIPS (median)\hspace*{-1em} & LPIPS-$\Delta$\\
  \midrule
  1-by-1 GAN projection  (all frames) & $0.33 {\pm} 0.10$ & $0.36$ & $0.154$ \\
  \midrule
  Separable kernel interp.\ (all f.) & $0.41 {\pm} 0.12$ & $0.42$ &  $0.026$ \\
  Quaternion kernel interp.\ (all f.) & $0.41 {\pm} 0.12$ & $0.43$ &  $\bm{0.021}$ \\
  GP interpolation (all frames) & $\bm{0.39} {\pm} \bm{0.13}$ & $\bm{0.41}$ & ${0.031}$ \\
  \midrule
  Linear interp.\ (first--last only) & $0.45 {\pm} 0.07$ & $0.46$ &  ${0.020}$ \\
  Separable kernel interp.\ (f--l) & $0.44 {\pm} 0.07$ & $0.46$ &  $0.024$ \\
  Quaternion kernel interp.\ (f--l) & $0.45 {\pm} 0.10$ & $\bm{0.44}$ &  $\bm{0.012}$ \\
  GP interp.\ (first--last only) & $\bm{0.42} {\pm} \bm{0.08}$ & $\bm{0.44}$ &  $0.020$ \\
  \bottomrule
  \end{tabular*} 
  \vspace*{-1em}
\end{table}

\paragraph{View synthesis} Next, we take only a subset of the frames---the extreme case with only a single start and a single end frame (see
\cref{fig:interpolation_comp})---and interpolate the rest of the frames in the latent space by predicting the latent codes, $\mathrm{E}[\vf(P_\star)\mid \mathcal{D}]$, for unseen views $P_\star$, following the {\em correlation structure of the original camera movement}.
In \cref{fig:interpolation_comp}, we compare to independent frame-by-frame reconstructions. For certain input head poses, the quality is gapped by suboptimal StyleGAN projections (leading to some variation in face alignment). As a baseline, we also linearly interpolate between the first and last frame, which (for apparent reasons) fails to capture the varying camera motion, with mismatches in the head angle. The GP solution with our view prior smoothly matches the view orientation while maintaining the face features. Also, we visualize the frame-wise marginal uncertainty (posterior variance $\mathrm{V}[\vf(P_j) \mid \mathcal{D}]$) of the GP predictions as a standard deviation map in image space. We create the maps by drawing 100 samples from the posterior process and calculating the standard deviation over faces. The uncertainty is small in the beginning/end (where the inputs are) and highest towards the part where the linear interpolation has the largest error---showing the practical uncertainty quantification capabilities of the model. We also measure the differences to ground-truth images (LPIPS in \cref{tbl:lpips}). One expects the direct StyleGAN projection that uses all frames to yield the minimum LPIPS, but it has poor temporal consistency (LPIPS-$\Delta$). The separable and quaternion kernels have it {\it vice versa}: Their high consistency (low LPIPS-$\Delta$) is irrelevant as it is due to losing the original diversity (increasing direct LPIPS, visuals in the supplement).
The start and end frames were selected for reasonable symmetry to fairly compare to linear interpolation. Still, the GP interpolation is clearly superior to the linear case. As expected, although GP interpolation with all frames reduces jitter (see supplementary video), it has less frame-by-frame similarity to the originals than direct projection.

\section{Discussion and Conclusion}
\label{sec:discussion}
We have presented a new GP covariance function to encode {\it a~priori} knowledge about camera movement into computer vision tasks, advocating more principled approaches for inter-frame reasoning in computer vision. We consider this view kernel an important building block for applying Gaussian process priors to many computer vision models. The covariance function itself is simple, yet elegant, and circumvents possible problems related to degeneracy and gimbal lock related to the alternative approaches. The model directly generalizes the standard periodic covariance function to high-dimensional rotations, filling a tangible gap in the existing GP tool set.

To underline the practical importance of our work, we considered real-world applications for the proposed model. Our quantitative experiments in \cref{sec:gppvae,sec:track} showed that the view prior can encode authentic movement and provide a soft-prior for view synthesis. We also showed (\cref{sec:face}) how the model can be of direct practical value by acting as a camera-motion-aware interpolator. Combining probabilistic models with computer vision tasks come with a promise of better data efficiency (not everything needs to be learned from data, as demonstrated in the comparison in \cref{sec:gppvae}) and uncertainty quantification (as in \cref{sec:face}).

\clearpage

\section*{Ethical Impact}

Following the breakthroughs of deep neural networks in recent years, broader societal concerns have increasingly shifted from maximizing the accuracy under controlled conditions to aspects such as robustness and explainability. In real-world applications, machine learning systems are expected to generalize despite limited amount of training data, yield principled quantification of uncertainty, and allow for human interpretation of the inference process.

Probabilistic methods provide natural solutions to these requirements. Yet, current Bayesian deep learning approaches fall short of ways to encode {\em interpretable} priors into models, in which non-parametric priors such as Gaussian processes can help.
These tools are widely used in, for instance, finance, navigation, and medical tasks, while computer vision applications have seen less benefit. Our work offers a principled building block that extends the gold standard Gaussian process tooling to allow utilization of Gaussian process priors across a range of computer vision tasks, of which we show-case just a few representative examples. We hope this work inspires computer vision practitioners of a variety of different subdomains to increasingly integrate probabilistic methods in their work, as well as motivate the researchers in probabilistic methods to explore models in computer vision applications.

\section*{Acknowledgments}
We thank Jaakko Lehtinen and Janne Hellsten (NVIDIA) for the StyleGAN latent space projection script and advice on its usage. Authors acknowledge funding from GenMind Ltd. and Academy of Finland (grant numbers 324345 and 308640). We acknowledge the computational resources provided by the Aalto Science-IT project.

\bibliography{bibliography}

\begin{thebibliography}{49}
\providecommand{\natexlab}[1]{#1}
\providecommand{\url}[1]{\texttt{#1}}
\providecommand{\urlprefix}{URL }
\expandafter\ifx\csname urlstyle\endcsname\relax
  \providecommand{\doi}[1]{doi:\discretionary{}{}{}#1}\else
  \providecommand{\doi}{doi:\discretionary{}{}{}\begingroup
  \urlstyle{rm}\Url}\fi

\bibitem[{Abdal, Qin, and Wonka(2019)}]{abdal:2019}
Abdal, R.; Qin, Y.; and Wonka, P. 2019.
\newblock {Image2StyleGAN}: {H}ow to embed images into the {StyleGAN} latent
  space?
\newblock In \emph{International Conference on Computer Vision (ICCV)},
  4432--4441.

\bibitem[{{Anderson} and {Barfoot}(2015)}]{7353368}
{Anderson}, S.; and {Barfoot}, T.~D. 2015.
\newblock Full STEAM ahead: {E}xactly sparse gaussian process regression for
  batch continuous-time trajectory estimation on {SE(3)}.
\newblock In \emph{IEEE/RSJ International Conference on Intelligent Robots and
  Systems (IROS)}, 157--164.

\bibitem[{Blundell et~al.(2015)Blundell, Cornebise, Kavukcuoglu, and
  Wierstra}]{Blundell+Cornebise+Kavukcuoglu+Wierstra:2015}
Blundell, C.; Cornebise, J.; Kavukcuoglu, K.; and Wierstra, D. 2015.
\newblock Weight uncertainty in neural network.
\newblock In \emph{Proceedings of the 32nd International Conference on Machine
  Learning (ICML)}, Proceedings of Machine Learning Research, 1613--1622. PMLR.

\bibitem[{Bui et~al.(2016)Bui, Hern{\'a}ndez-Lobato, Hernandez-Lobato, Li, and
  Turner}]{Bui:2016}
Bui, T.; Hern{\'a}ndez-Lobato, D.; Hernandez-Lobato, J.; Li, Y.; and Turner, R.
  2016.
\newblock Deep {G}aussian processes for regression using approximate
  expectation propagation.
\newblock In \emph{Proceedings of the 32nd International Conference on Machine
  Learning (ICML)}, Proceedings of Machine Learning Research, 1472--1481. PMLR.

\bibitem[{Casale et~al.(2018)Casale, Dalca, Saglietti, Listgarten, and
  Fusi}]{Casale+Dalca+Saglietti+Listgarten+Fusi:2018}
Casale, F.~P.; Dalca, A.; Saglietti, L.; Listgarten, J.; and Fusi, N. 2018.
\newblock Gaussian process prior variational autoencoders.
\newblock In \emph{Advances in Neural Information Processing Systems
  (NeurIPS)}, 10369--10380. Curran Associates, Inc.

\bibitem[{Chang et~al.(2015)Chang, Funkhouser, Guibas, Hanrahan, Huang, Li,
  Savarese, Savva, Song, Su, Xiao, Yi, and
  Yu}]{Chang+Funkhouser+Guibas+Hanrahan+Huang+Li+Savarese+Savva+Song+Su+others:2015}
Chang, A.~X.; Funkhouser, T.~A.; Guibas, L.~J.; Hanrahan, P.; Huang, Q.-X.; Li,
  Z.; Savarese, S.; Savva, M.; Song, S.; Su, H.; Xiao, J.; Yi, L.; and Yu, F.
  2015.
\newblock {ShapeNet}: {A}n information-rich {3D} model repository.
\newblock \emph{arXiv preprint arXiv:1512.03012} .

\bibitem[{Cort{\'e}s et~al.(2018)Cort{\'e}s, Solin, Rahtu, and
  Kannala}]{Cortes+Solin+Rahtu+Kannala:2018}
Cort{\'e}s, S.; Solin, A.; Rahtu, E.; and Kannala, J. 2018.
\newblock {ADVIO:} {A}n authentic dataset for visual-inertial odometry.
\newblock In \emph{Proceedings of the European Conference on Computer Vision
  (ECCV)}, 419--434.

\bibitem[{Deisenroth and Rasmussen(2011)}]{Deisenroth:2011}
Deisenroth, M.~P.; and Rasmussen, C.~E. 2011.
\newblock {PILCO:} {A} model-based and data-efficient approach to policy
  search.
\newblock In \emph{Proceedings of the 28th International Conference on Machine
  Learning (ICML)}, 465--472. Omnipress.

\bibitem[{Diebel(2006)}]{Diebel:2006}
Diebel, J. 2006.
\newblock Representing attitude: {E}uler angles, unit quaternions, and rotation
  vectors.
\newblock \emph{Matrix} 58(15-16): 1--35.

\bibitem[{Dutordoir et~al.(2020)Dutordoir, {van der Wilk}, Artemev, Tomczak,
  and Hensman}]{Dutordoir+Wilk+Artemev+Tomczak+Hensman:2019}
Dutordoir, V.; {van der Wilk}, M.; Artemev, A.; Tomczak, M.; and Hensman, J.
  2020.
\newblock Translation insensitivity for deep convolutional {G}aussian
  processes.
\newblock In \emph{International Conference on Artificial Intelligence and
  Statistics (AISTATS)}.

\bibitem[{Duvenaud(2014)}]{Duvenaud:2014}
Duvenaud, D. 2014.
\newblock \emph{Automatic Model Construction with {G}aussian Processes}.
\newblock Ph.D. thesis, Computational and Biological Learning Laboratory,
  University of Cambridge, Cambridge, UK.

\bibitem[{Eleftheriadis et~al.(2016)Eleftheriadis, Rudovic, Deisenroth, and
  Pantic}]{Eleftheriadis+Rudovic+Deisenroth+Pantic:2016}
Eleftheriadis, S.; Rudovic, O.; Deisenroth, M.~P.; and Pantic, M. 2016.
\newblock Variational {G}aussian process auto-encoder for ordinal prediction of
  facial action units.
\newblock In \emph{Asian Conference on Computer Vision (ACCV)}, 154--170.
  Springer.

\bibitem[{Eleftheriadis, Rudovic, and
  Pantic(2015{\natexlab{a}})}]{Eleftheriadis+Rudovic+Pantic:2015a}
Eleftheriadis, S.; Rudovic, O.; and Pantic, M. 2015{\natexlab{a}}.
\newblock Discriminative shared {G}aussian processes for multiview and
  view-invariant facial expression recognition.
\newblock \emph{IEEE Transactions on Image Processing} 24(1): 189--204.

\bibitem[{Eleftheriadis, Rudovic, and
  Pantic(2015{\natexlab{b}})}]{Eleftheriadis+Rudovic+Pantic:2015b}
Eleftheriadis, S.; Rudovic, O.; and Pantic, M. 2015{\natexlab{b}}.
\newblock Multi-conditional latent variable model for joint facial action unit
  detection.
\newblock In \emph{IEEE International Conference on Computer Vision (ICCV)},
  3792--3800.

\bibitem[{Everingham, Sivic, and
  Zisserman(2006)}]{Everingham+Sivic+Zisserman:2006}
Everingham, M.; Sivic, J.; and Zisserman, A. 2006.
\newblock ``{H}ello! {M}y name is... {B}uffy''--{A}utomatic naming of
  characters in {TV} video.
\newblock In \emph{British Machine Vision Conference (BMVC)}.

\bibitem[{Featherstone(2014)}]{Featherstone:2014}
Featherstone, R. 2014.
\newblock \emph{Rigid Body Dynamics Algorithms}.
\newblock New York: Springer.

\bibitem[{Gardner et~al.(2018)Gardner, Pleiss, Weinberger, Bindel, and
  Wilson}]{Gardner+Pleiss+Weinberger+Bindel+Wilson:2018}
Gardner, J.; Pleiss, G.; Weinberger, K.~Q.; Bindel, D.; and Wilson, A.~G. 2018.
\newblock {GPyTorch}: Blackbox matrix-matrix {G}aussian process inference with
  {GPU} acceleration.
\newblock In \emph{Advances in Neural Information Processing Systems
  (NeurIPS)}, 7576--7586. Curran Associates, Inc.

\bibitem[{Goodfellow et~al.(2014)Goodfellow, Pouget-Abadie, Mirza, Xu,
  Warde-Farley, Ozair, Courville, and
  Bengio}]{Goodfellow+Pouget-Abadie+Mirza+Xu+Warde-Farley+Ozair+Courville+Bengio:2014}
Goodfellow, I.; Pouget-Abadie, J.; Mirza, M.; Xu, B.; Warde-Farley, D.; Ozair,
  S.; Courville, A.; and Bengio, Y. 2014.
\newblock Generative adversarial nets.
\newblock In \emph{Advances in Neural Information Processing Systems (NIPS)},
  2672--2680.

\bibitem[{Haasdonk and Burkhardt(2007)}]{Haasdonk+Burkhardt:2007}
Haasdonk, B.; and Burkhardt, H. 2007.
\newblock Invariant kernel functions for pattern analysis and machine learning.
\newblock \emph{Machine Learning} 68(1): 35--61.

\bibitem[{Hamsici and Martinez(2008)}]{hamsici2008rotation}
Hamsici, O.~C.; and Martinez, A.~M. 2008.
\newblock Rotation invariant kernels and their application to shape analysis.
\newblock \emph{IEEE Transactions on Pattern Analysis and Machine Intelligence}
  31(11): 1985--1999.

\bibitem[{Hartley and Zisserman(2003)}]{Hartley+Zisserman:2003}
Hartley, R.; and Zisserman, A. 2003.
\newblock \emph{Multiple View Geometry in Computer Vision}.
\newblock Cambridge University Press.

\bibitem[{Hennig, Osborne, and Girolami(2015)}]{Hennig:2015}
Hennig, P.; Osborne, M.~A.; and Girolami, M. 2015.
\newblock Probabilistic numerics and uncertainty in computations.
\newblock \emph{Proceedings of the Royal Society of London {A:} Mathematical,
  Physical and Engineering Sciences} 471(2179).

\bibitem[{Hensman, Durrande, and Solin(2018)}]{Hensman+Durrande+Solin:2018}
Hensman, J.; Durrande, N.; and Solin, A. 2018.
\newblock Variational {F}ourier features for {G}aussian processes.
\newblock \emph{Journal of Machine Learning Research (JMLR)} 18(151): 1–52.

\bibitem[{Hensman, Fusi, and Lawrence(2013)}]{Hensman+Fusi+Lawrence:2013}
Hensman, J.; Fusi, N.; and Lawrence, N.~D. 2013.
\newblock Gaussian processes for big data.
\newblock In \emph{Uncertainty in Artificial Intelligence (UAI)}, 282--290.
  AUAI Press.

\bibitem[{Hou et~al.(2021)Hou, Janjua, Kannala, and
  Solin}]{Hou+Janjua+Kannala+Solin:2021}
Hou, Y.; Janjua, M.~K.; Kannala, J.; and Solin, A. 2021.
\newblock Movement-induced Priors for Deep Stereo.
\newblock In \emph{International Conference on Pattern Recognition (ICPR)}.

\bibitem[{Hou, Kannala, and Solin(2019)}]{Hou+Kannala+Solin:2019}
Hou, Y.; Kannala, J.; and Solin, A. 2019.
\newblock Multi-view stereo by temporal nonparametric fusion.
\newblock In \emph{IEEE International Conference on Computer Vision (ICCV)},
  2651--2660.

\bibitem[{Jayasumana et~al.(2013)Jayasumana, Hartley, Salzmann, Li, and
  Harandi}]{Jayasumana:2013}
Jayasumana, S.; Hartley, R.; Salzmann, M.; Li, H.; and Harandi, M. 2013.
\newblock Kernel methods on the {R}iemannian manifold of symmetric positive
  definite matrices.
\newblock In \emph{IEEE Conference on Computer Vision and Pattern Recognition
  (CVPR)}, 73--80.

\bibitem[{Karras, Laine, and Aila(2019)}]{Karras+Laine+Aila:2019}
Karras, T.; Laine, S.; and Aila, T. 2019.
\newblock A style-based generator architecture for generative adversarial
  networks.
\newblock In \emph{IEEE Conference on Computer Vision and Pattern Recognition
  (CVPR)}, 4401--4410.

\bibitem[{Kazemi and Sullivan(2015)}]{kazemi:2015}
Kazemi, V.; and Sullivan, J. 2015.
\newblock One millisecond face alignment with an ensemble of regression trees.
\newblock In \emph{IEEE Conference on Computer Vision and Pattern Recognition
  (CVPR)}, 1867--1874.

\bibitem[{Kendall and Gal(2017)}]{Kendall+Gal:2017}
Kendall, A.; and Gal, Y. 2017.
\newblock What uncertainties do we need in {B}ayesian deep learning for
  computer vision?
\newblock In \emph{Advances in Neural Information Processing Systems (NIPS)},
  5574--5584.

\bibitem[{Krauth et~al.(2017)Krauth, Bonilla, Cutajar, and
  Filippone}]{Krauth+Bonilla+Cutajar+Filippone:2017}
Krauth, K.; Bonilla, E.~V.; Cutajar, K.; and Filippone, M. 2017.
\newblock {AutoGP}: {E}xploring the capabilities and limitations of {G}aussian
  process models.
\newblock In \emph{Uncertainty in Artificial Intelligence (UAI)}. AUAI Press.

\bibitem[{Lang and Hirche(2017)}]{Lang+Hirche:2017}
Lang, M.; and Hirche, S. 2017.
\newblock Computationally efficient rigid-body Gaussian process for motion
  dynamics.
\newblock \emph{IEEE Robotics and Automation Letters} 2(3): 1601--1608.

\bibitem[{Lang, Kleinsteuber, and Hirche(2018)}]{Lang+Kleinsteuber+Hirche:2018}
Lang, M.; Kleinsteuber, M.; and Hirche, S. 2018.
\newblock Gaussian process for {6-DoF} rigid motions.
\newblock \emph{Autonomous Robots} 42(6): 1151--1167.

\bibitem[{L{\'a}zaro-Gredilla et~al.(2010)L{\'a}zaro-Gredilla,
  Qui{\~n}onero-Candela, Rasmussen, and
  Figueiras-Vidal}]{Lazaro-Gredilla+Quinonero-Candela+Rasmussen+Figueiras-Vidal:2010}
L{\'a}zaro-Gredilla, M.; Qui{\~n}onero-Candela, J.; Rasmussen, C.~E.; and
  Figueiras-Vidal, A.~R. 2010.
\newblock Sparse spectrum {G}aussian process regression.
\newblock \emph{Journal of Machine Learning Research (JMLR)} 11: 1865--1881.

\bibitem[{Lucas and Kanade(1981)}]{Lucas+Kanade:1981}
Lucas, B.~D.; and Kanade, T. 1981.
\newblock An iterative image registration technique with an application to
  stereo vision.
\newblock In \emph{International Conference on Artificial Intelligence
  (IJCAI)}, 674--679. Vancouver, BC, Canada.

\bibitem[{MacKay(1998)}]{MacKay:1998}
MacKay, D.~J. 1998.
\newblock Introduction to {G}aussian processes.
\newblock \emph{NATO ASI Series F Computer and Systems Sciences} 168: 133--166.

\bibitem[{Matthews et~al.(2017)Matthews, {van der Wilk}, Nickson, Fujii,
  Boukouvalas, Le{\'o}n-Villagr{\'a}, Ghahramani, and Hensman}]{Matthews:2017}
Matthews, A. G. d.~G.; {van der Wilk}, M.; Nickson, T.; Fujii, K.; Boukouvalas,
  A.; Le{\'o}n-Villagr{\'a}, P.; Ghahramani, Z.; and Hensman, J. 2017.
\newblock {GPflow}: {A} {G}aussian process library using {TensorFlow}.
\newblock \emph{Journal of Machine Learning Research (JMLR)} 18(1): 1299--1304.

\bibitem[{Mazzotti, Sancisi, and
  Parenti-Castelli(2016)}]{Mazzotti+Sancisi+Parenti-Castelli:2016}
Mazzotti, C.; Sancisi, N.; and Parenti-Castelli, V. 2016.
\newblock A measure of the distance between two rigid-body poses based on the
  use of platonic solids.
\newblock In \emph{ROMANSY 21-Robot Design, Dynamics and Control}, 81--89.
  Springer.

\bibitem[{{Puzer (GitHub user)}(2019)}]{puzer:2019}
{Puzer (GitHub user)}. 2019.
\newblock {StyleGAN} Encoder -- {C}onverts real images to latent space.
\newblock \url{https://github.com/Puzer/stylegan-encoder}.
\newblock {GitHub} repository.

\bibitem[{Rasmussen and Williams(2006)}]{Rasmussen+Williams:2006}
Rasmussen, C.~E.; and Williams, C. K.~I. 2006.
\newblock \emph{{G}aussian Processes for Machine Learning}.
\newblock The {MIT} {P}ress.

\bibitem[{S\"arkk\"a, Solin, and
  Hartikainen(2013)}]{Sarkka+Solin+Hartikainen:2013}
S\"arkk\"a, S.; Solin, A.; and Hartikainen, J. 2013.
\newblock Spatiotemporal learning via infinite-dimensional {B}ayesian filtering
  and smoothing.
\newblock \emph{IEEE Signal Processing Magazine} 30(4): 51--61.

\bibitem[{Shi and Tomasi(1994)}]{Shi+Tomasi:1994}
Shi, J.; and Tomasi, C. 1994.
\newblock Good features to track.
\newblock In \emph{IEEE Conference on Computer Vision and Pattern Recognition
  (CVPR)}, 593--600.

\bibitem[{Simonyan and Zisserman(2015)}]{simonyan:2015}
Simonyan, K.; and Zisserman, A. 2015.
\newblock Very deep convolutional networks for large-scale image recognition.
\newblock In \emph{International Conference on Learning Representations
  (ICLR)}.

\bibitem[{Solin, Hensman, and Turner(2018)}]{Solin+Hensman+Turner:2018}
Solin, A.; Hensman, J.; and Turner, R.~E. 2018.
\newblock Infinite-horizon {G}aussian processes.
\newblock In \emph{Advances in Neural Information Processing Systems
  (NeurIPS)}, 3486--3495. Curran Associates, Inc.

\bibitem[{Song et~al.(2009)Song, Huang, Smola, and Fukumizu}]{song2009hilbert}
Song, L.; Huang, J.; Smola, A.; and Fukumizu, K. 2009.
\newblock Hilbert space embeddings of conditional distributions with
  applications to dynamical systems.
\newblock In \emph{Proceedings of the 26th Annual International Conference on
  Machine Learning (ICML)}, 961--968.

\bibitem[{Urtasun, Fleet, and Fua(2006)}]{Urtasun:2006}
Urtasun, R.; Fleet, D.~J.; and Fua, P. 2006.
\newblock {3D} people tracking with {G}aussian process dynamical models.
\newblock In \emph{IEEE Conference on Computer Vision and Pattern Recognition
  (CVPR)}, 238--245.

\bibitem[{Wang et~al.(2019)Wang, Pleiss, Gardner, Tyree, Weinberger, and
  Wilson}]{Wang+Pleiss+Gardner+Tyree+Weinberger+Wilson:2019}
Wang, K.~A.; Pleiss, G.; Gardner, J.~R.; Tyree, S.; Weinberger, K.~Q.; and
  Wilson, A.~G. 2019.
\newblock Exact {G}aussian processes on a million data points.
\newblock In \emph{Advances in Neural Information Processing Systems
  (NeurIPS)}, 14622--14632. Curran Associates, Inc.

\bibitem[{Wilson and Nickisch(2015)}]{Wilson+Nickisch:2015-ICML}
Wilson, A.~G.; and Nickisch, H. 2015.
\newblock Kernel interpolation for scalable structured {G}aussian processes
  ({KISS-GP}).
\newblock In \emph{International Conference on Machine Learning (ICML)},
  volume~37 of \emph{PMLR}, 1775--1784.

\bibitem[{Zhang et~al.(2018)Zhang, Isola, Efros, Shechtman, and
  Wang}]{zhang2018}
Zhang, R.; Isola, P.; Efros, A.~A.; Shechtman, E.; and Wang, O. 2018.
\newblock The unreasonable effectiveness of deep features as a perceptual
  metric.
\newblock In \emph{Proceedings of the IEEE Conference on Computer Vision and
  Pattern Recognition (CVPR)}, 586--595.

\end{thebibliography}

\clearpage

\twocolumn[%
  \begin{center}
  \Large\bf Supplementary material for\\
  Gaussian Process Priors for View-Aware Inference
  \end{center}
  \vspace*{2em}
]

\appendix\setcounter{section}{0}\pagestyle{empty}

\medskip

\noindent
This file holds the supplementary material for `Gaussian Process Priors for View-Aware Inference'. We provide further details on the derivations, experiment setups, additional results, and details on the video supplement.

\section{Derivations}

\subsection{Derivation of the Geodesic Distance}
\label{app:geodesic}
The geodesic arc distance along the $\mathrm{S}^3$ sphere defined by a rotation $\MR \in \mathrm{SO}(3)$ can be intuitively derived from the following linear algebraic and trigonometric identities. Consider the eigendecomposition of $\MR$ where the eigenvectors define the rotation axis and the eigenvalues define the angle (or `distance'). The eigenvalues of a rotation matrix come as a complex conjugate pair and `1'. This can be represented as $\{1, e^{\pm\mathrm{i}\theta}\}$, where $\theta$ is the rotation angle or arc distance.

In order to solve $\theta$ given $\MR$, recall the definition of cosine in terms of complex numbers:
\begin{equation}
  \cos \theta = \frac{1}{2} \left(e^{\mathrm{i}\theta} + e^{-\mathrm{i}\theta} \right).
\end{equation}

The trace of a matrix equals the sum of its eigenvalues, and using the definition of cosine form above, we recover:
\begin{equation}
  \mathrm{tr}(\MR) = 1 + e^{\mathrm{i}\theta} + e^{-\mathrm{i}\theta} = 1 + 2 \cos \theta.
\end{equation}
Solving for $\theta$ gives
\begin{equation}
  \theta = \mathrm{arccos}\bigg(\frac{\mathrm{tr}(\MR)-1}{2}\bigg).
\end{equation}
The relative rotation between two orientations $\MR_1, \MR_2 \in \mathrm{SO}(3)$ is given by $\MR_1\T\MR_2$, thus we recover the geodesic distance metric that was used in \cref{sec:view-kernels} in the main paper:
\begin{equation}\label{eq:d-geo}
  d_\mathrm{g}(\MR_1, \MR_2) = \mathrm{arccos}\bigg(\frac{\mathrm{tr}(\MR_1\T \MR_2)-1}{2}\bigg).
\end{equation}

\subsection{Derivation of the Three-Dimensional View Kernel}
\label{app:deriv}
We seek to derive the 3D counterpart of the standard periodic kernel. For this we set up a local approximation around the origin for the geodesic distance. Recall that the Taylor series of cosine:
\begin{equation}
  \cos \theta = 1 - \frac{1}{2}\theta^2 + \frac{1}{4!}\theta^4 + \cdots
\end{equation}
Truncating after the second term gives $\cos \theta \approx 1 - \frac{1}{2}\theta^2$. Substituting $\sqrt{2\alpha}$ for $\theta$ yields $\cos \sqrt{2\alpha} \approx 1 - \alpha$. Thus we get the local approximation
\begin{equation}
  \mathrm{arccos}(1 - \alpha) \approx \sqrt{2\alpha}.
\end{equation}

Applying the above to the geodesic distance in \cref{eq:d-geo}, and substituting $\alpha$ for $\frac{1}{2}(3 - \mathrm{tr}(\MR_1\T\MR_2))$, we get 
\begin{align}
  d_\mathrm{g}(\MR_1, \MR_2) &\approx \sqrt{2\alpha} 
  = \sqrt{3 - \mathrm{tr}(\MR_1\T\MR_2)} 
  = \sqrt{\mathrm{tr}(\MI - \MR_1\T\MR_2)},
\end{align}
which we label the `view' distance
\begin{equation}\label{eq:d-view-supp}
  d_\mathrm{view}(\MR_1, \MR_2) = \sqrt{\mathrm{tr}(\MI - \MR_1\T \MR_2)}.
\end{equation}
As was seen in the main paper, this approximation is accurate around origin (see \cref{fig:distance-e}).

\subsection{The Link Between the Standard Periodic and the 3D View Kernel}
\label{app:link}
We define a general rotation matrix $\MR$ with respect to yaw, pitch, and roll defined through rotations around the axes, $x$, $y$, and $z$, respectively:
\begin{align}
  \MR_\mathrm{x}(\theta ) &=
  \begin{pmatrix}
    1 & 0 & 0\\
    0 & \cos \theta & -\sin \theta \\
    0 & \sin \theta & \cos \theta
  \end{pmatrix}, \\
  \MR_\mathrm{y}(\theta ) &= 
  \begin{pmatrix}
    \cos \theta & 0 & \sin \theta \\
    0 & 1 & 0 \\
    -\sin \theta & 0 & \cos \theta 
  \end{pmatrix}, \text{~and}\\
  \MR_\mathrm{z}(\theta ) &= 
  \begin{pmatrix}
    \cos \theta & -\sin \theta & 0 \\
    \sin \theta &\cos \theta & 0 \\
    0&0&1
  \end{pmatrix}.
\end{align}
Thus a rotation around the axes can be given by Euler angles $\vtheta = (\theta_1,\theta_2,\theta_3)$: $\MR(\vtheta) = \MR_\mathrm{z}(\theta_3)\,\MR_\mathrm{y}(\theta_2)\,\MR_\mathrm{x}(\theta_1)$.

Taking the trace term from \cref{eq:d-view-supp}, we can parametrize the rotations in terms of $\vtheta$ and $\vtheta'$, giving:
\begin{equation}
  \mathrm{tr}(\MI - \MR(\vtheta)\T \MR(\vtheta')).
\end{equation}
In the degenerate case where there is only rotation around one of the axes (say $\theta_1 = \theta_1'$ and $\theta_2 = \theta_2'$,  $\theta_3 = \theta$ and $\theta_3' = \theta'$), we get
\begin{equation}
  \mathrm{tr}(\MI - \MR_\mathrm{z}(\theta)\T \MR_\mathrm{z}(\theta')).
\end{equation}
Expanding the product and summing up the diagonal elements gives
\begin{equation}
  (1 - \cos(\theta-\theta')) + (1 - \cos(\theta-\theta')) + 0 = 2 - 2\cos(\theta-\theta').
\end{equation}
By applying the half-angle formula ($\sin^2(\frac{1}{2}\alpha) = \frac{1}{2}(1-\cos \alpha)$), we recover
\begin{equation}
  4 \sin^2\left( \frac{\theta-\theta'}{2} \right),
\end{equation}
which is exactly the form of the squared norm warping in the standard periodic covariance function \cite{MacKay:1998,Rasmussen+Williams:2006}: $\|\vu(\theta)-\vu(\theta')\|^2$ for $\vu(\theta) = (\cos \theta, \sin \theta)$. The above derivation can be repeated for any of the angles $\{\theta_1, \theta_2, \theta_3\}$, showing that in the case of rotation only around one axis, the proposed view covariance function coincides with the standard periodic covariance function.

\begin{figure*}[!t]
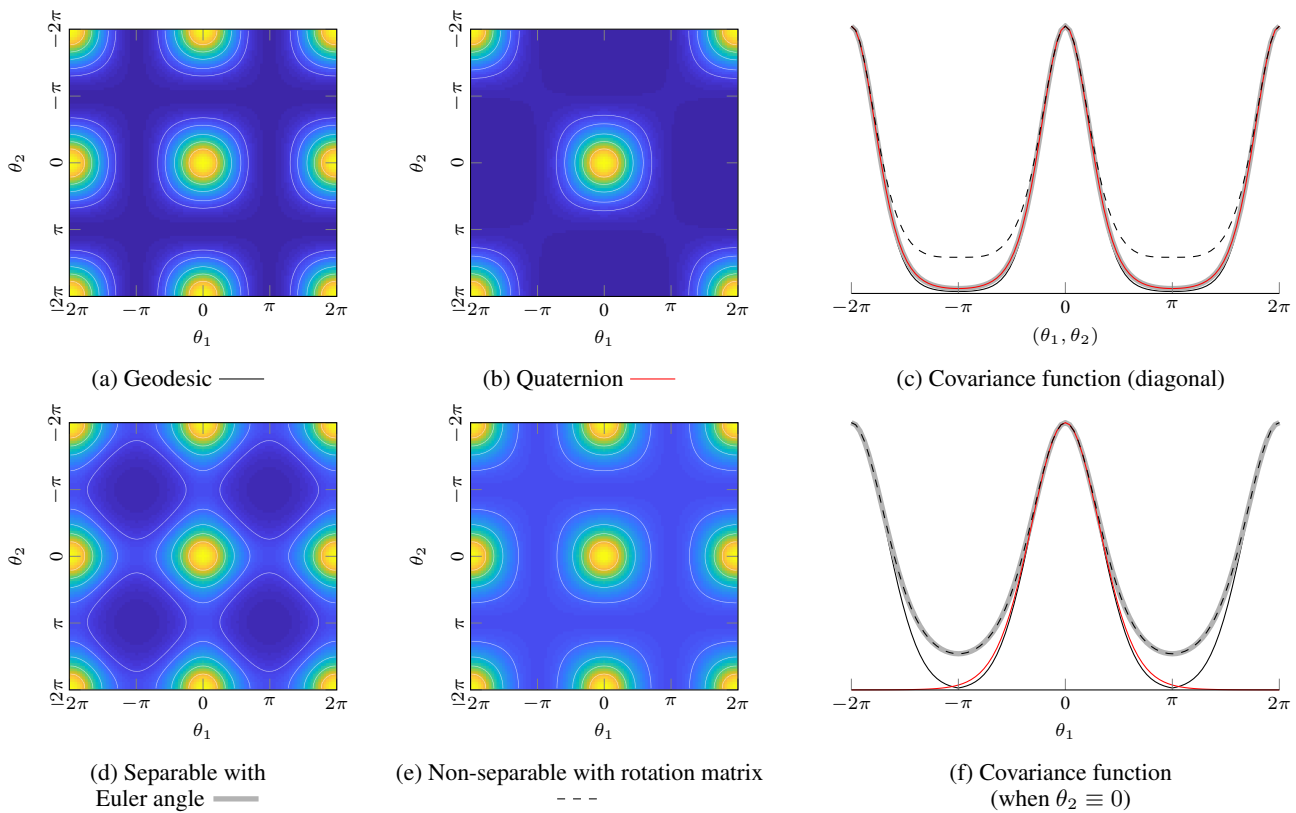

  \centering\scriptsize
  \pgfplotsset{yticklabel style={rotate=90}, ylabel style={yshift=-15pt},clip=true,scale only axis,axis on top,clip marker paths,legend style={row sep=0pt},legend columns=-1,xlabel near ticks,clip=false}
  \setlength{\figurewidth}{.2\textwidth}
  \setlength{\figureheight}{\figurewidth}
  \captionsetup[subfigure]{justification=centering}
  \begin{subfigure}[b]{.28\textwidth}
    \centering
%
%
\begin{tikzpicture}

\begin{axis}[%
point meta min=0,
point meta max=1,
axis on top,
xmin=-6.29235784047474,
xmax=6.29235784047474,
xtick={-6.28318530717959,-3.14159265358979,0,3.14159265358979,6.28318530717959},
xticklabels={{$-2\pi$},{$-\pi$},{0},{$\pi$},{$2\pi$}},
xlabel={$\theta_1$},
y dir=reverse,
ymin=-6.29234446943495,
ymax=6.29234446943495,
ytick={-6.28318530717959,-3.14159265358979,0,3.14159265358979,6.28318530717959},
yticklabels={{$-2\pi$},{$-\pi$},{0},{$\pi$},{$2\pi$}},
ylabel={$\theta_2$},
axis background/.style={fill=white},
legend style={legend cell align=left,align=left,draw=white!15!black},
width=\figurewidth,
height=\figureheight
]
\addplot [forget plot] graphics [xmin=-6.29235784047474,xmax=6.29235784047474,ymin=-6.29234446943495,ymax=6.29234446943495] {./fig/cov-geodesic-1.png};
\end{axis}
\end{tikzpicture}%
    \caption{Geodesic \ref{addplot:distance0}}
    \label{fig:cov-a}
  \end{subfigure}
  \hspace*{\fill}
  \begin{subfigure}[b]{.28\textwidth}
    \centering
%
%
\begin{tikzpicture}

\begin{axis}[%
point meta min=0,
point meta max=1,
axis on top,
xmin=-6.29235784047474,
xmax=6.29235784047474,
xtick={-6.28318530717959,-3.14159265358979,0,3.14159265358979,6.28318530717959},
xticklabels={{$-2\pi$},{$-\pi$},{0},{$\pi$},{$2\pi$}},
xlabel={$\theta_1$},
y dir=reverse,
ymin=-6.29234446943495,
ymax=6.29234446943495,
ytick={-6.28318530717959,-3.14159265358979,0,3.14159265358979,6.28318530717959},
yticklabels={{$-2\pi$},{$-\pi$},{0},{$\pi$},{$2\pi$}},
ylabel={$\theta_2$},
axis background/.style={fill=white},
legend style={legend cell align=left,align=left,draw=white!15!black},
width=\figurewidth,
height=\figureheight
]
\addplot [forget plot] graphics [xmin=-6.29235784047474,xmax=6.29235784047474,ymin=-6.29234446943495,ymax=6.29234446943495] {./fig/cov-quaternion-1.png};
\end{axis}
\end{tikzpicture}%
    \caption{Quaternion \ref{addplot:distance2}}
    \label{fig:cov-b}    
  \end{subfigure}  
  \hspace*{\fill}
  \begin{subfigure}[b]{.4\textwidth}
    \setlength{\figureheight}{\figurewidth}  
    \setlength{\figurewidth}{.8\textwidth}  
    \pgfplotsset{y axis line style={draw opacity=0}}
    \centering
    \input{./fig/diag-cov.tex}
    \caption{Covariance function (diagonal)}
    \label{fig:cov-c}    
  \end{subfigure}\\
  \begin{subfigure}[b]{.28\textwidth}
    \centering
%
%
\begin{tikzpicture}

\begin{axis}[%
point meta min=0,
point meta max=1,
axis on top,
xmin=-6.29235784047474,
xmax=6.29235784047474,
xtick={-6.28318530717959,-3.14159265358979,0,3.14159265358979,6.28318530717959},
xticklabels={{$-2\pi$},{$-\pi$},{0},{$\pi$},{$2\pi$}},
xlabel={$\theta_1$},
y dir=reverse,
ymin=-6.29234446943495,
ymax=6.29234446943495,
ytick={-6.28318530717959,-3.14159265358979,0,3.14159265358979,6.28318530717959},
yticklabels={{$-2\pi$},{$-\pi$},{0},{$\pi$},{$2\pi$}},
ylabel={$\theta_2$},
axis background/.style={fill=white},
legend style={legend cell align=left,align=left,draw=white!15!black},
width=\figurewidth,
height=\figureheight
]
\addplot [forget plot] graphics [xmin=-6.29235784047474,xmax=6.29235784047474,ymin=-6.29234446943495,ymax=6.29234446943495] {./fig/cov-separable-1.png};
\end{axis}
\end{tikzpicture}%
    \caption{Separable with \\ Euler angle \ref{addplot:distance1}}
    \label{fig:cov-d}    
  \end{subfigure}
  \hspace*{\fill}
  \begin{subfigure}[b]{.28\textwidth}
    \centering
%
%
\begin{tikzpicture}

\begin{axis}[%
point meta min=0,
point meta max=1,
axis on top,
xmin=-6.29235784047474,
xmax=6.29235784047474,
xtick={-6.28318530717959,-3.14159265358979,0,3.14159265358979,6.28318530717959},
xticklabels={{$-2\pi$},{$-\pi$},{0},{$\pi$},{$2\pi$}},
xlabel={$\theta_1$},
y dir=reverse,
ymin=-6.29234446943495,
ymax=6.29234446943495,
ytick={-6.28318530717959,-3.14159265358979,0,3.14159265358979,6.28318530717959},
yticklabels={{$-2\pi$},{$-\pi$},{0},{$\pi$},{$2\pi$}},
ylabel={$\theta_2$},
axis background/.style={fill=white},
legend style={legend cell align=left,align=left,draw=white!15!black},
width=\figurewidth,
height=\figureheight
]
\addplot [forget plot] graphics [xmin=-6.29235784047474,xmax=6.29235784047474,ymin=-6.29234446943495,ymax=6.29234446943495] {./fig/cov-trace-1.png};
\end{axis}
\end{tikzpicture}%
    \caption{Non-separable with rotation matrix \ref{addplot:distance3}}
    \label{fig:cov-e}    
  \end{subfigure}    
  \hspace*{\fill}
  \begin{subfigure}[b]{.4\textwidth}
    \setlength{\figureheight}{\figurewidth}  
    \setlength{\figurewidth}{.8\textwidth}
    \pgfplotsset{y axis line style={draw opacity=0}}    
    \centering
    \input{./fig/covariance.tex}
    \caption{Covariance function \\(when $\theta_2 \equiv 0$)}
    \label{fig:cov-f}    
  \end{subfigure}
  \caption{{\bf (Left)}~Covariance function between two degrees-of-freedom rotations (for simpler visualization) with scale $0$~\protect\includegraphics[width=1cm]{fig/parula}~$1$ and $\ell=1$. (a)~uses the geodesic distance (see \cref{eq:d-rot}), (b)~the quaternion norm distance (see \cref{eq:d-quat-norm}), (d)~shows the separable periodic covariance function with Euler angle (see \cref{eq:per-sep}), and (e)~the proposed non-separable  covariance function. {\bf (Right)}~Cross-sections along the diagonal and $\theta_2 \equiv 0$, showing that in 1D (d) and (e) coincide, while (e) is symmetric in 2D/3D.}
  \label{fig:covariances}
  \vspace*{-1em}  
\end{figure*}

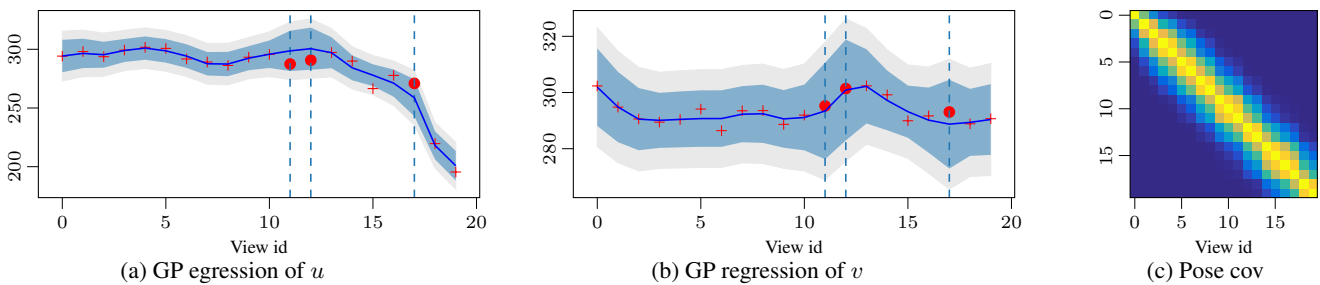
\begin{figure*}[!t]
  \centering\scriptsize
  \pgfplotsset{yticklabel style={rotate=90}, ylabel style={yshift=-15pt},clip=true,scale only axis,axis on top,clip marker paths,legend style={row sep=0pt},legend columns=-1,xlabel near ticks,clip=false}
  \setlength{\figurewidth}{.14\textwidth}
  \setlength{\figureheight}{\figurewidth}
  \begin{subfigure}[b]{.33\textwidth}
    \setlength{\figureheight}{\figurewidth}  
    \setlength{\figurewidth}{\textwidth}
    \centering
\begin{tikzpicture}

\definecolor{color0}{rgb}{0.12156862745098,0.466666666666667,0.705882352941177}

\begin{axis}[
tick align=outside,
tick pos=left,
x grid style={white!69.01960784313725!black},
xlabel={View id},
xmin=-1.15600358422939, xmax=20.1560035842294,
xtick style={color=black},
y grid style={white!69.01960784313725!black},
ymin=173.654876887833, ymax=333.015273691882,
ytick style={color=black},
width=\figurewidth,
height=\figureheight
]
\path [draw=white!82.74509803921568!black, fill=white!82.74509803921568!black, opacity=0.5]
(axis cs:0,273.124269716084)
--(axis cs:0,315.378726977764)
--(axis cs:1,316.348066169006)
--(axis cs:2,313.759203751507)
--(axis cs:3,316.123705478773)
--(axis cs:4,318.272615553391)
--(axis cs:5,316.047435800064)
--(axis cs:6,311.590653596437)
--(axis cs:7,304.984259852809)
--(axis cs:8,305.11552368321)
--(axis cs:9,309.997270168297)
--(axis cs:10,314.826390860764)
--(axis cs:11,323.127746286239)
--(axis cs:12,325.771619291698)
--(axis cs:13,317.639110569313)
--(axis cs:14,302.160312333673)
--(axis cs:15,294.539322821218)
--(axis cs:16,290.065287561357)
--(axis cs:17,281.875181799407)
--(axis cs:18,237.402334084807)
--(axis cs:19,220.737113312979)
--(axis cs:19,180.898531288017)
--(axis cs:19,180.898531288017)
--(axis cs:18,198.89778294988)
--(axis cs:17,235.6300844862)
--(axis cs:16,252.257376674443)
--(axis cs:15,261.322670869188)
--(axis cs:14,266.247610963285)
--(axis cs:13,276.552839780542)
--(axis cs:12,275.240580024668)
--(axis cs:11,274.15510742221)
--(axis cs:10,276.76085916805)
--(axis cs:9,274.830669980541)
--(axis cs:8,269.754340953624)
--(axis cs:7,270.363057949117)
--(axis cs:6,276.836255315884)
--(axis cs:5,281.296752854082)
--(axis cs:4,283.738399347077)
--(axis cs:3,281.877995055187)
--(axis cs:2,276.977408051)
--(axis cs:1,276.664209630504)
--(axis cs:0,273.124269716084)
--cycle;

\path [fill=color0, fill opacity=0.5]
(axis cs:0,280.449876178648)
--(axis cs:0,308.0531205152)
--(axis cs:1,309.022459706442)
--(axis cs:2,306.433597288943)
--(axis cs:3,308.798099016209)
--(axis cs:4,310.947009090827)
--(axis cs:5,308.7218293375)
--(axis cs:6,304.265047133873)
--(axis cs:7,297.658653390245)
--(axis cs:8,297.789917220646)
--(axis cs:9,302.671663705734)
--(axis cs:10,307.5007843982)
--(axis cs:11,315.802139823675)
--(axis cs:12,318.446012829134)
--(axis cs:13,310.313504106749)
--(axis cs:14,294.83470587111)
--(axis cs:15,287.213716358654)
--(axis cs:16,282.739681098793)
--(axis cs:17,274.549575336843)
--(axis cs:18,230.076727622243)
--(axis cs:19,213.411506850415)
--(axis cs:19,188.224137750581)
--(axis cs:19,188.224137750581)
--(axis cs:18,206.223389412443)
--(axis cs:17,242.955690948763)
--(axis cs:16,259.582983137006)
--(axis cs:15,268.648277331752)
--(axis cs:14,273.573217425848)
--(axis cs:13,283.878446243105)
--(axis cs:12,282.566186487232)
--(axis cs:11,281.480713884774)
--(axis cs:10,284.086465630614)
--(axis cs:9,282.156276443104)
--(axis cs:8,277.079947416188)
--(axis cs:7,277.68866441168)
--(axis cs:6,284.161861778448)
--(axis cs:5,288.622359316646)
--(axis cs:4,291.064005809641)
--(axis cs:3,289.203601517751)
--(axis cs:2,284.303014513564)
--(axis cs:1,283.989816093068)
--(axis cs:0,280.449876178648)
--cycle;

\addplot [only marks, mark = +, draw=red, fill=red, colormap/viridis]
table{%
x                      y
0 294.18
1 298.07
2 293.7
3 299.31
4 301.69
5 300.76
6 291.75
7 289.18
8 286.21
9 293.25
10 295.64
13 297.3
14 289.96
15 266.53
16 277.76
18 219.65
19 195.48
};
\addplot [only marks, draw=red, fill=red, colormap/viridis]
table{%
x                      y
11 287.42
12 290.71
17 270.94
};
\addplot [semithick, color0, dashed]
table {%
11 173.654876887833
11 333.015273691882
};
\addplot [semithick, color0, dashed]
table {%
12 173.654876887833
12 333.015273691882
};
\addplot [semithick, color0, dashed]
table {%
17 173.654876887833
17 333.015273691882
};
\addplot [semithick, blue]
table {%
0 294.251498346924
1 296.506137899755
2 295.368305901254
3 299.00085026698
4 301.005507450234
5 298.672094327073
6 294.213454456161
7 287.673658900963
8 287.434932318417
9 292.413970074419
10 295.793625014407
11 298.641426854225
12 300.506099658183
13 297.095975174927
14 284.203961648479
15 277.930996845203
16 271.1613321179
17 258.752633142803
18 218.150058517343
19 200.817822300498
};
\end{axis}

\end{tikzpicture}\\[-6pt]
    \caption{GP egression of $u$ }
  \end{subfigure}
  \hspace*{\fill}
  \begin{subfigure}[b]{.33\textwidth}
    \setlength{\figureheight}{\figurewidth}  
    \setlength{\figurewidth}{\textwidth}
    \centering
\begin{tikzpicture}

\definecolor{color0}{rgb}{0.12156862745098,0.466666666666667,0.705882352941177}

\begin{axis}[
tick align=outside,
tick pos=left,
x grid style={white!69.01960784313725!black},
xlabel={View id},
xmin=-1.15600358422939, xmax=20.1560035842294,
xtick style={color=black},
y grid style={white!69.01960784313725!black},
ymin=262.592371879685, ymax=329.173911286602,
ytick style={color=black},
width=\figurewidth,
height=\figureheight
]
\path [draw=white!82.74509803921568!black, fill=white!82.74509803921568!black, opacity=0.5]
(axis cs:0,280.784486778419)
--(axis cs:0,323.038944040099)
--(axis cs:1,314.684145197146)
--(axis cs:2,308.917929259255)
--(axis cs:3,307.210800099739)
--(axis cs:4,307.70233711329)
--(axis cs:5,308.093291677379)
--(axis cs:6,308.107925227162)
--(axis cs:7,309.592242797763)
--(axis cs:8,310.108265456174)
--(axis cs:9,308.192797039766)
--(axis cs:10,310.141331907033)
--(axis cs:11,317.903717433652)
--(axis cs:12,326.147477677197)
--(axis cs:13,322.744745762797)
--(axis cs:14,315.140895152413)
--(axis cs:15,309.790700249558)
--(axis cs:16,309.070641309923)
--(axis cs:17,311.863902802298)
--(axis cs:18,308.621079609096)
--(axis cs:19,310.319351622781)
--(axis cs:19,270.48076959782)
--(axis cs:19,270.48076959782)
--(axis cs:18,270.116528474169)
--(axis cs:17,265.618805489091)
--(axis cs:16,271.262730423008)
--(axis cs:15,276.574048297528)
--(axis cs:14,279.228193782024)
--(axis cs:13,281.658474974025)
--(axis cs:12,275.616438410167)
--(axis cs:11,268.931078569624)
--(axis cs:10,272.07580021432)
--(axis cs:9,273.02619685201)
--(axis cs:8,274.747082726588)
--(axis cs:7,274.971040894071)
--(axis cs:6,273.353526946609)
--(axis cs:5,273.342608731398)
--(axis cs:4,273.168120906976)
--(axis cs:3,272.965089676154)
--(axis cs:2,272.136133558748)
--(axis cs:1,275.000288658645)
--(axis cs:0,280.784486778419)
--cycle;

\path [fill=color0, fill opacity=0.5]
(axis cs:0,288.110093240983)
--(axis cs:0,315.713337577535)
--(axis cs:1,307.358538734583)
--(axis cs:2,301.592322796691)
--(axis cs:3,299.885193637175)
--(axis cs:4,300.376730650726)
--(axis cs:5,300.767685214815)
--(axis cs:6,300.782318764599)
--(axis cs:7,302.266636335199)
--(axis cs:8,302.78265899361)
--(axis cs:9,300.867190577203)
--(axis cs:10,302.81572544447)
--(axis cs:11,310.578110971088)
--(axis cs:12,318.821871214633)
--(axis cs:13,315.419139300233)
--(axis cs:14,307.815288689849)
--(axis cs:15,302.465093786994)
--(axis cs:16,301.745034847359)
--(axis cs:17,304.538296339734)
--(axis cs:18,301.295473146533)
--(axis cs:19,302.993745160217)
--(axis cs:19,277.806376060384)
--(axis cs:19,277.806376060384)
--(axis cs:18,277.442134936732)
--(axis cs:17,272.944411951654)
--(axis cs:16,278.588336885572)
--(axis cs:15,283.899654760092)
--(axis cs:14,286.553800244588)
--(axis cs:13,288.984081436589)
--(axis cs:12,282.942044872731)
--(axis cs:11,276.256685032187)
--(axis cs:10,279.401406676883)
--(axis cs:9,280.351803314573)
--(axis cs:8,282.072689189151)
--(axis cs:7,282.296647356635)
--(axis cs:6,280.679133409173)
--(axis cs:5,280.668215193962)
--(axis cs:4,280.493727369539)
--(axis cs:3,280.290696138718)
--(axis cs:2,279.461740021312)
--(axis cs:1,282.325895121208)
--(axis cs:0,288.110093240983)
--cycle;

\addplot [only marks, mark = +, draw=red, fill=red, colormap/viridis]
table{%
x                      y
0 302.34
1 294.8
2 290.57
3 289.45
4 290.37
5 294.11
6 286.43
7 293.5
8 293.56
9 288.67
10 291.95
13 302.37
14 299.18
15 289.95
16 291.67
18 288.85
19 290.68
};
\addplot [only marks, draw=red, fill=red, colormap/viridis]
table{%
x                      y
11 295.22
12 301.39
17 293.07
};
\addplot [semithick, color0, dashed]
table {%
11 262.592371879685
11 329.173911286602
};
\addplot [semithick, color0, dashed]
table {%
12 262.592371879685
12 329.173911286602
};
\addplot [semithick, color0, dashed]
table {%
17 262.592371879685
17 329.173911286602
};
\addplot [semithick, blue]
table {%
0 301.911715409259
1 294.842216927896
2 290.527031409001
3 290.087944887946
4 290.435229010133
5 290.717950204389
6 290.730726086886
7 292.281641845917
8 292.42767409138
9 290.609496945888
10 291.108566060677
11 293.417398001638
12 300.881958043682
13 302.201610368411
14 297.184544467218
15 293.182374273543
16 290.166685866466
17 288.741354145694
18 289.368804041633
19 290.4000606103
};
\end{axis}

\end{tikzpicture}\\[-6pt]
    \caption{GP regression of $v$ }
  \end{subfigure}  
  \hspace*{\fill}
  \begin{subfigure}[b]{.2\textwidth}
    \centering
\begin{tikzpicture}

\begin{axis}[
tick align=outside,
tick pos=left,
x grid style={white!69.01960784313725!black},
xmin=-0.5, xmax=19.5,
xtick style={color=black},
xlabel={View id},
y grid style={white!69.01960784313725!black},
ymin=-0.5, ymax=19.5,
ytick style={color=black},
y dir = reverse,
width=\figurewidth,
height=\figureheight
]
\addplot graphics [includegraphics cmd=\pgfimage,xmin=-0.5, xmax=19.5, ymin=19.5, ymax=-0.5] {./fig/tracker/002-K-000.png};
\end{axis}

\end{tikzpicture}\\[-6pt]
    \caption{Pose cov}
  \end{subfigure}
  \caption{Experiment for seeing how the view-aware kernel can capture authentic camera movement. GP regression for one example feature track using the pose kernel. The red points corresponds to ground-truth trajectories, where `\ref{addplot:marker-plus}' means training points and `\ref{addplot:marker-bullet}' corresponds to unseen points. The blue points are predicted mean values. The shaded patches denote the 95\% posterior and posterior predictive quantiles.}
  \label{fig:track-example}
\end{figure*}

\section{Comparison of Camera Motion Kernels}
\label{sec:track}
Apart from showing how to use the proposed view kernel for inter-frame reasoning, we compare the performance of different kernels in visual feature track tasks, evaluating their flexibility and capability with real-world camera motion.

The data \citep[from][]{Cortes+Solin+Rahtu+Kannala:2018} contains handheld motion of a Google Tango device while walking in a shopping mall. It comprises a video with 789 frames (resolution $640{\times}480$) and associated camera pose information for every frame. We apply a Shi--Tomasi feature extractor \cite{Shi+Tomasi:1994} to select strong corner points in the frames and track them across frames by a pyramidal Lucas--Kanade tracker \cite{Lucas+Kanade:1981}. 
We discard short tracks and are left with 533 full-length feature tracks, tracked over 20~frames each, giving input--output pairs $\{(P_i,(u,v)_i)\}_{i=1}^{20}$, where $(u,v)$ are the pixel coordinates and $P_i$ denotes the camera pose. In each track, we use 85\% of  points for training and 15\% for testing.

\begin{figure}[!t]
  \centering\small
  \pgfplotsset{yticklabel style={rotate=90}, ylabel style={yshift=-15pt},scale only axis,axis on top,clip=false, xlabel near ticks}
  \setlength{\figurewidth}{.4\textwidth}
  \setlength{\figureheight}{.55\figurewidth}  
  \begin{minipage}{\columnwidth}
  \captionof{table}{RMSE and NLPD measures for the feature tracking experiment.
 Results calculated over 533 tracks of handheld movement. {\bf Smaller is better.} \label{tbl:results}}  
  \small
  \begin{tabular*}{\textwidth}{@{\extracolsep{\fill}} lcccc}
  \toprule
  Model (kernel) description & RMSE & NLPD \\
  \midrule

  Linear-in-extrinsics \cite{Lang+Hirche:2017} & 9.56 & 3.692\\
  Translation only & 8.01 & 3.679 \\

  Trans.\ \& geodesic & 7.58 & 3.616\\
  Trans.\ \& quaternion & 7.67 & 3.621\\
  Trans.\ \& sep.\ orientation & 7.54 & \textbf{3.438}\\
  Trans.\ \& non-sep. orientation (ours) & \textbf{7.44} & 3.591\\
  
  \bottomrule
  \end{tabular*} 
  \end{minipage}  
  \vspace*{-2em}
\end{figure}

We set up a GP regression task for predicting the $u$ and $v$ coordinates for unseen frames along the track. As the world coordinates points are regarded fixed, the point locations in the frames are driven by the movement of the camera (\cf, \cref{eq:camera}). We consider independent GP regression models for every track $j=1,2,\ldots,533$ in $u$ and $v$: $u_i^{(j)} = f_{\mathrm{u}}^{(j)}(P_i^{(j)}) + \varepsilon_{\mathrm{u},i}^{(j)}$ and $v_i^{(j)} = f_{\mathrm{v}}^{(j)}(P_i^{(j)}) + \varepsilon_{\mathrm{v},i}^{(j)}$ with GP priors $f_{\mathrm{u}}^{(j)} \sim \mathrm{GP}(0,\kappa(P,P'))$ and $f_{\mathrm{v}}^{(j)} \sim \mathrm{GP}(0,\kappa(P,P'))$. \cref{fig:track-example} shows an example of the task.
In contrast to the other two experiments, here the views depend on both rotation and relatively large translational movement. We compare six different movement-induced kernels: {\em (i)}~a linear-in-extrinsics kernel \citep[see the dot-product kernel in][]{Lang+Hirche:2017}, {\em (ii)}~only translation (\cref{eq:trans}), and {\em (iii--vi)}~product kernels between the translation kernel and each orientation kernel. For each covariance function,we jointly learn the hyperparameters of the GP priors by maximizing w.r.t.\ log likelihood.

We evaluate the models on the test data in terms of predictive RMSE and negative log-predictive density (NLPD). The results in \cref{tbl:results} show the proposed non-separable view orientation kernel outperforms all other orientation kernels on RMSE. In this experiment, the camera has a lot of forward movement, causing translation-only model to already achieve a relatively large improvement. All stationary kernels---even the translation-only model with the standard periodic kernel---outperform the linear dot product kernel that uses the full 6-DoF pose. All results with orientation kernels show that considering orientation covariance provides a clear benefit. Though our proposed non-separable kernel is derived from geodesic distance, it outperforms the geodesic kernel due to also accounting for different characteristic scaling per axis, which makes it readily non-isotropic. Both the separable kernel with Euler angles and our proposed non-separable kernel with rotation matrix can be regarded as extensions of the 1D standard periodic kernel and they both perform well in the task. However, due to the well-known deficiencies of Euler angles, the separable view kernel should be avoided in practical applications.

\begin{figure*}[!t]
  \centering\scriptsize
  \pgfplotsset{yticklabel style={rotate=90}, ylabel style={yshift=-15pt},clip=true,scale only axis,axis on top,clip marker paths,legend style={row sep=0pt},legend columns=-1,xlabel near ticks,clip=false}
  \captionsetup[subfigure]{justification=centering}
  \setlength{\figurewidth}{.17\textwidth}
  \setlength{\figureheight}{\figurewidth}
  \begin{subfigure}[b]{.34\textwidth}
    \setlength{\figureheight}{\figurewidth}  
    \setlength{\figurewidth}{.93\textwidth}
    \centering
\begin{tikzpicture}

\definecolor{color0}{rgb}{0.12156862745098,0.466666666666667,0.705882352941177}

\begin{axis}[
tick align=outside,
tick pos=left,
x grid style={white!69.01960784313725!black},
xlabel={View id},
xmin=-1.15600358422939, xmax=20.1560035842294,
xtick style={color=black},
y grid style={white!69.01960784313725!black},
ymin=482.747981946768, ymax=581.198288280331,
ytick style={color=black},
width=\figurewidth,
height=\figureheight
]
\path [draw=white!82.74509803921568!black, fill=white!82.74509803921568!black, opacity=0.5]
(axis cs:0,487.22299587102)
--(axis cs:0,521.693150643206)
--(axis cs:1,521.361139158142)
--(axis cs:2,522.796959879934)
--(axis cs:3,525.296909845533)
--(axis cs:4,530.940057998577)
--(axis cs:5,536.795960764415)
--(axis cs:6,537.357544203119)
--(axis cs:7,537.799783776934)
--(axis cs:8,539.588054744615)
--(axis cs:9,540.622934328208)
--(axis cs:10,542.055407876531)
--(axis cs:11,543.71756293404)
--(axis cs:12,550.414438582559)
--(axis cs:13,552.381170228114)
--(axis cs:14,554.678485736093)
--(axis cs:15,557.653181026738)
--(axis cs:16,561.067756123966)
--(axis cs:17,564.385574449547)
--(axis cs:18,573.683540235497)
--(axis cs:19,576.723274356078)
--(axis cs:19,535.58354344266)
--(axis cs:19,535.58354344266)
--(axis cs:18,535.980966447212)
--(axis cs:17,532.613292468999)
--(axis cs:16,530.985331588411)
--(axis cs:15,529.09287336499)
--(axis cs:14,526.899004683493)
--(axis cs:13,524.584687403492)
--(axis cs:12,522.044960058822)
--(axis cs:11,514.740004029345)
--(axis cs:10,513.614255734588)
--(axis cs:9,512.538145233848)
--(axis cs:8,511.399081743638)
--(axis cs:7,507.103916889223)
--(axis cs:6,505.529801825906)
--(axis cs:5,503.970343573578)
--(axis cs:4,498.6407648993)
--(axis cs:3,496.219363239987)
--(axis cs:2,494.581290630404)
--(axis cs:1,491.677079789306)
--(axis cs:0,487.22299587102)
--cycle;

\path [fill=color0, fill opacity=0.5]
(axis cs:0,494.548602333584)
--(axis cs:0,514.367544180642)
--(axis cs:1,514.035532695579)
--(axis cs:2,515.47135341737)
--(axis cs:3,517.97130338297)
--(axis cs:4,523.614451536014)
--(axis cs:5,529.470354301851)
--(axis cs:6,530.031937740555)
--(axis cs:7,530.47417731437)
--(axis cs:8,532.262448282051)
--(axis cs:9,533.297327865645)
--(axis cs:10,534.729801413967)
--(axis cs:11,536.391956471476)
--(axis cs:12,543.088832119995)
--(axis cs:13,545.055563765551)
--(axis cs:14,547.352879273529)
--(axis cs:15,550.327574564174)
--(axis cs:16,553.742149661402)
--(axis cs:17,557.059967986983)
--(axis cs:18,566.357933772934)
--(axis cs:19,569.397667893515)
--(axis cs:19,542.909149905224)
--(axis cs:19,542.909149905224)
--(axis cs:18,543.306572909775)
--(axis cs:17,539.938898931563)
--(axis cs:16,538.310938050975)
--(axis cs:15,536.418479827553)
--(axis cs:14,534.224611146057)
--(axis cs:13,531.910293866056)
--(axis cs:12,529.370566521386)
--(axis cs:11,522.065610491909)
--(axis cs:10,520.939862197151)
--(axis cs:9,519.863751696412)
--(axis cs:8,518.724688206202)
--(axis cs:7,514.429523351787)
--(axis cs:6,512.85540828847)
--(axis cs:5,511.295950036142)
--(axis cs:4,505.966371361864)
--(axis cs:3,503.544969702551)
--(axis cs:2,501.906897092968)
--(axis cs:1,499.00268625187)
--(axis cs:0,494.548602333584)
--cycle;

\addplot [only marks, mark = +, draw=red, fill=red, colormap/viridis]
table{%
x                      y
0 500.85
1 505.49
2 509.46
3 514.46
4 515.26
6 519.78
7 521.72
8 524.24
9 527.07
10 529.15
11 531.69
12 534.49
13 537.38
14 539.51
15 544
16 547.24
19 557.57
};
\addplot [only marks, draw=red, fill=red, colormap/viridis]
table{%
x                      y
5 516.94
17 550.63
18 554.6
};
\addplot [semithick, color0, dashed]
table {%
5 482.747981946768
5 581.198288280331
};
\addplot [semithick, color0, dashed]
table {%
17 482.747981946768
17 581.198288280331
};
\addplot [semithick, color0, dashed]
table {%
18 482.747981946768
18 581.198288280331
};
\addplot [semithick, blue]
table {%
0 504.458073257113
1 506.519109473724
2 508.689125255169
3 510.75813654276
4 514.790411448939
5 520.383152168996
6 521.443673014513
7 522.451850333079
8 525.493568244126
9 526.580539781028
10 527.834831805559
11 529.228783481692
12 536.229699320691
13 538.482928815803
14 540.788745209793
15 543.373027195864
16 546.026543856188
17 548.499433459273
18 554.832253341355
19 556.153408899369
};
\end{axis}

\end{tikzpicture}
  \end{subfigure}
  \hspace*{\fill}
  \begin{subfigure}[b]{.34\textwidth}
    \setlength{\figureheight}{\figurewidth}  
    \setlength{\figurewidth}{.93\textwidth}
    \centering
\begin{tikzpicture}

\definecolor{color0}{rgb}{0.12156862745098,0.466666666666667,0.705882352941177}

\begin{axis}[
tick align=outside,
tick pos=left,
x grid style={white!69.01960784313725!black},
xlabel={View id},
xmin=-1.15600358422939, xmax=20.1560035842294,
xtick style={color=black},
y grid style={white!69.01960784313725!black},
ymin=125.174350948029, ymax=241.148997113522,
ytick style={color=black},
width=\figurewidth,
height=\figureheight
]
\path [draw=white!82.74509803921568!black, fill=white!82.74509803921568!black, opacity=0.5]
(axis cs:0,201.407267515632)
--(axis cs:0,235.877422287817)
--(axis cs:1,233.054807072304)
--(axis cs:2,231.739042536711)
--(axis cs:3,231.524973268185)
--(axis cs:4,231.744719831075)
--(axis cs:5,230.820795817052)
--(axis cs:6,230.484948707067)
--(axis cs:7,230.202474272102)
--(axis cs:8,229.222554943733)
--(axis cs:9,228.553816253934)
--(axis cs:10,227.371607211844)
--(axis cs:11,225.377582036274)
--(axis cs:12,207.184102612625)
--(axis cs:13,200.482982825152)
--(axis cs:14,194.088179010572)
--(axis cs:15,187.674505278146)
--(axis cs:16,181.992669639933)
--(axis cs:17,177.439562990997)
--(axis cs:18,170.441783623038)
--(axis cs:19,171.585656687151)
--(axis cs:19,130.445925773733)
--(axis cs:19,130.445925773733)
--(axis cs:18,132.739209834752)
--(axis cs:17,145.66728101045)
--(axis cs:16,151.910245104378)
--(axis cs:15,159.114197616398)
--(axis cs:14,166.308697957973)
--(axis cs:13,172.68650000053)
--(axis cs:12,178.814624088888)
--(axis cs:11,196.400023131579)
--(axis cs:10,198.930455069901)
--(axis cs:9,200.469027159574)
--(axis cs:8,201.033581942756)
--(axis cs:7,199.506607384392)
--(axis cs:6,198.657206329855)
--(axis cs:5,197.995178626215)
--(axis cs:4,199.445426731798)
--(axis cs:3,202.447426662639)
--(axis cs:2,203.523373287181)
--(axis cs:1,203.370747703468)
--(axis cs:0,201.407267515632)
--cycle;

\path [fill=color0, fill opacity=0.5]
(axis cs:0,208.732873978196)
--(axis cs:0,228.551815825254)
--(axis cs:1,225.72920060974)
--(axis cs:2,224.413436074147)
--(axis cs:3,224.199366805621)
--(axis cs:4,224.419113368511)
--(axis cs:5,223.495189354488)
--(axis cs:6,223.159342244503)
--(axis cs:7,222.876867809539)
--(axis cs:8,221.896948481169)
--(axis cs:9,221.228209791371)
--(axis cs:10,220.04600074928)
--(axis cs:11,218.05197557371)
--(axis cs:12,199.858496150061)
--(axis cs:13,193.157376362588)
--(axis cs:14,186.762572548008)
--(axis cs:15,180.348898815583)
--(axis cs:16,174.667063177369)
--(axis cs:17,170.113956528434)
--(axis cs:18,163.116177160474)
--(axis cs:19,164.260050224587)
--(axis cs:19,137.771532236297)
--(axis cs:19,137.771532236297)
--(axis cs:18,140.064816297316)
--(axis cs:17,152.992887473014)
--(axis cs:16,159.235851566942)
--(axis cs:15,166.439804078962)
--(axis cs:14,173.634304420536)
--(axis cs:13,180.012106463093)
--(axis cs:12,186.140230551452)
--(axis cs:11,203.725629594143)
--(axis cs:10,206.256061532465)
--(axis cs:9,207.794633622138)
--(axis cs:8,208.35918840532)
--(axis cs:7,206.832213846956)
--(axis cs:6,205.982812792419)
--(axis cs:5,205.320785088779)
--(axis cs:4,206.771033194362)
--(axis cs:3,209.773033125203)
--(axis cs:2,210.848979749745)
--(axis cs:1,210.696354166032)
--(axis cs:0,208.732873978196)
--cycle;

\addplot [only marks, mark = +, draw=red, fill=red, colormap/viridis]
table{%
x                      y
0 219.46
1 220.12
2 216.56
3 215.46
4 215.6
6 216.44
7 214.64
8 214.12
9 214.89
10 212.75
11 207.39
12 200.11
13 190.99
14 182.65
15 167.43
16 160.4
19 151.79
};
\addplot [only marks, draw=red, fill=red, colormap/viridis]
table{%
x                      y
5 216.66
17 156.95
18 154.36
};
\addplot [semithick, color0, dashed]
table {%
5 125.174350948029
5 241.148997113522
};
\addplot [semithick, color0, dashed]
table {%
17 125.174350948029
17 241.148997113522
};
\addplot [semithick, color0, dashed]
table {%
18 125.174350948029
18 241.148997113522
};
\addplot [semithick, blue]
table {%
0 218.642344901725
1 218.212777387886
2 217.631207911946
3 216.986199965412
4 215.595073281437
5 214.407987221634
6 214.571077518461
7 214.854540828247
8 215.128068443244
9 214.511421706754
10 213.151031140872
11 210.888802583926
12 192.999363350757
13 186.584741412841
14 180.198438484272
15 173.394351447272
16 166.951457372156
17 161.553422000724
18 151.590496728895
19 151.015791230442
};
\end{axis}

\end{tikzpicture}
  \end{subfigure}  
  \hspace*{\fill}
  \begin{subfigure}[b]{.2\textwidth}
    \centering
\begin{tikzpicture}

\begin{axis}[
tick align=outside,
tick pos=left,
x grid style={white!69.01960784313725!black},
xmin=-0.5, xmax=19.5,
xtick style={color=black},
xlabel={View id},
y grid style={white!69.01960784313725!black},
ymin=-0.5, ymax=19.5,
ytick style={color=black},
y dir = reverse,
width=\figurewidth,
height=\figureheight
]
\addplot graphics [includegraphics cmd=\pgfimage,xmin=-0.5, xmax=19.5, ymin=19.5, ymax=-0.5] {./fig/tracker/001-K-000.png};
\end{axis}

\end{tikzpicture}  
  \end{subfigure}\\
  \begin{subfigure}[b]{.34\textwidth}
    \setlength{\figureheight}{\figurewidth}  
    \setlength{\figurewidth}{.93\textwidth}
    \centering
\begin{tikzpicture}

\definecolor{color0}{rgb}{0.12156862745098,0.466666666666667,0.705882352941177}

\begin{axis}[
tick align=outside,
tick pos=left,
x grid style={white!69.01960784313725!black},
xlabel={View id},
xmin=-1.15600358422939, xmax=20.1560035842294,
xtick style={color=black},
y grid style={white!69.01960784313725!black},
ymin=173.654876887833, ymax=333.015273691882,
ytick style={color=black},
width=\figurewidth,
height=\figureheight
]
\path [draw=white!82.74509803921568!black, fill=white!82.74509803921568!black, opacity=0.5]
(axis cs:0,273.124269716084)
--(axis cs:0,315.378726977764)
--(axis cs:1,316.348066169006)
--(axis cs:2,313.759203751507)
--(axis cs:3,316.123705478773)
--(axis cs:4,318.272615553391)
--(axis cs:5,316.047435800064)
--(axis cs:6,311.590653596437)
--(axis cs:7,304.984259852809)
--(axis cs:8,305.11552368321)
--(axis cs:9,309.997270168297)
--(axis cs:10,314.826390860764)
--(axis cs:11,323.127746286239)
--(axis cs:12,325.771619291698)
--(axis cs:13,317.639110569313)
--(axis cs:14,302.160312333673)
--(axis cs:15,294.539322821218)
--(axis cs:16,290.065287561357)
--(axis cs:17,281.875181799407)
--(axis cs:18,237.402334084807)
--(axis cs:19,220.737113312979)
--(axis cs:19,180.898531288017)
--(axis cs:19,180.898531288017)
--(axis cs:18,198.89778294988)
--(axis cs:17,235.6300844862)
--(axis cs:16,252.257376674443)
--(axis cs:15,261.322670869188)
--(axis cs:14,266.247610963285)
--(axis cs:13,276.552839780542)
--(axis cs:12,275.240580024668)
--(axis cs:11,274.15510742221)
--(axis cs:10,276.76085916805)
--(axis cs:9,274.830669980541)
--(axis cs:8,269.754340953624)
--(axis cs:7,270.363057949117)
--(axis cs:6,276.836255315884)
--(axis cs:5,281.296752854082)
--(axis cs:4,283.738399347077)
--(axis cs:3,281.877995055187)
--(axis cs:2,276.977408051)
--(axis cs:1,276.664209630504)
--(axis cs:0,273.124269716084)
--cycle;

\path [fill=color0, fill opacity=0.5]
(axis cs:0,280.449876178648)
--(axis cs:0,308.0531205152)
--(axis cs:1,309.022459706442)
--(axis cs:2,306.433597288943)
--(axis cs:3,308.798099016209)
--(axis cs:4,310.947009090827)
--(axis cs:5,308.7218293375)
--(axis cs:6,304.265047133873)
--(axis cs:7,297.658653390245)
--(axis cs:8,297.789917220646)
--(axis cs:9,302.671663705734)
--(axis cs:10,307.5007843982)
--(axis cs:11,315.802139823675)
--(axis cs:12,318.446012829134)
--(axis cs:13,310.313504106749)
--(axis cs:14,294.83470587111)
--(axis cs:15,287.213716358654)
--(axis cs:16,282.739681098793)
--(axis cs:17,274.549575336843)
--(axis cs:18,230.076727622243)
--(axis cs:19,213.411506850415)
--(axis cs:19,188.224137750581)
--(axis cs:19,188.224137750581)
--(axis cs:18,206.223389412443)
--(axis cs:17,242.955690948763)
--(axis cs:16,259.582983137006)
--(axis cs:15,268.648277331752)
--(axis cs:14,273.573217425848)
--(axis cs:13,283.878446243105)
--(axis cs:12,282.566186487232)
--(axis cs:11,281.480713884774)
--(axis cs:10,284.086465630614)
--(axis cs:9,282.156276443104)
--(axis cs:8,277.079947416188)
--(axis cs:7,277.68866441168)
--(axis cs:6,284.161861778448)
--(axis cs:5,288.622359316646)
--(axis cs:4,291.064005809641)
--(axis cs:3,289.203601517751)
--(axis cs:2,284.303014513564)
--(axis cs:1,283.989816093068)
--(axis cs:0,280.449876178648)
--cycle;

\addplot [only marks, mark = +, draw=red, fill=red, colormap/viridis]
table{%
x                      y
0 294.18
1 298.07
2 293.7
3 299.31
4 301.69
5 300.76
6 291.75
7 289.18
8 286.21
9 293.25
10 295.64
13 297.3
14 289.96
15 266.53
16 277.76
18 219.65
19 195.48
};
\addplot [only marks, draw=red, fill=red, colormap/viridis]
table{%
x                      y
11 287.42
12 290.71
17 270.94
};
\addplot [semithick, color0, dashed]
table {%
11 173.654876887833
11 333.015273691882
};
\addplot [semithick, color0, dashed]
table {%
12 173.654876887833
12 333.015273691882
};
\addplot [semithick, color0, dashed]
table {%
17 173.654876887833
17 333.015273691882
};
\addplot [semithick, blue]
table {%
0 294.251498346924
1 296.506137899755
2 295.368305901254
3 299.00085026698
4 301.005507450234
5 298.672094327073
6 294.213454456161
7 287.673658900963
8 287.434932318417
9 292.413970074419
10 295.793625014407
11 298.641426854225
12 300.506099658183
13 297.095975174927
14 284.203961648479
15 277.930996845203
16 271.1613321179
17 258.752633142803
18 218.150058517343
19 200.817822300498
};
\end{axis}

\end{tikzpicture}
  \end{subfigure}
  \hspace*{\fill}
  \begin{subfigure}[b]{.34\textwidth}
    \setlength{\figureheight}{\figurewidth}  
    \setlength{\figurewidth}{.93\textwidth}
    \centering
\begin{tikzpicture}

\definecolor{color0}{rgb}{0.12156862745098,0.466666666666667,0.705882352941177}

\begin{axis}[
tick align=outside,
tick pos=left,
x grid style={white!69.01960784313725!black},
xlabel={View id},
xmin=-1.15600358422939, xmax=20.1560035842294,
xtick style={color=black},
y grid style={white!69.01960784313725!black},
ymin=262.592371879685, ymax=329.173911286602,
ytick style={color=black},
width=\figurewidth,
height=\figureheight
]
\path [draw=white!82.74509803921568!black, fill=white!82.74509803921568!black, opacity=0.5]
(axis cs:0,280.784486778419)
--(axis cs:0,323.038944040099)
--(axis cs:1,314.684145197146)
--(axis cs:2,308.917929259255)
--(axis cs:3,307.210800099739)
--(axis cs:4,307.70233711329)
--(axis cs:5,308.093291677379)
--(axis cs:6,308.107925227162)
--(axis cs:7,309.592242797763)
--(axis cs:8,310.108265456174)
--(axis cs:9,308.192797039766)
--(axis cs:10,310.141331907033)
--(axis cs:11,317.903717433652)
--(axis cs:12,326.147477677197)
--(axis cs:13,322.744745762797)
--(axis cs:14,315.140895152413)
--(axis cs:15,309.790700249558)
--(axis cs:16,309.070641309923)
--(axis cs:17,311.863902802298)
--(axis cs:18,308.621079609096)
--(axis cs:19,310.319351622781)
--(axis cs:19,270.48076959782)
--(axis cs:19,270.48076959782)
--(axis cs:18,270.116528474169)
--(axis cs:17,265.618805489091)
--(axis cs:16,271.262730423008)
--(axis cs:15,276.574048297528)
--(axis cs:14,279.228193782024)
--(axis cs:13,281.658474974025)
--(axis cs:12,275.616438410167)
--(axis cs:11,268.931078569624)
--(axis cs:10,272.07580021432)
--(axis cs:9,273.02619685201)
--(axis cs:8,274.747082726588)
--(axis cs:7,274.971040894071)
--(axis cs:6,273.353526946609)
--(axis cs:5,273.342608731398)
--(axis cs:4,273.168120906976)
--(axis cs:3,272.965089676154)
--(axis cs:2,272.136133558748)
--(axis cs:1,275.000288658645)
--(axis cs:0,280.784486778419)
--cycle;

\path [fill=color0, fill opacity=0.5]
(axis cs:0,288.110093240983)
--(axis cs:0,315.713337577535)
--(axis cs:1,307.358538734583)
--(axis cs:2,301.592322796691)
--(axis cs:3,299.885193637175)
--(axis cs:4,300.376730650726)
--(axis cs:5,300.767685214815)
--(axis cs:6,300.782318764599)
--(axis cs:7,302.266636335199)
--(axis cs:8,302.78265899361)
--(axis cs:9,300.867190577203)
--(axis cs:10,302.81572544447)
--(axis cs:11,310.578110971088)
--(axis cs:12,318.821871214633)
--(axis cs:13,315.419139300233)
--(axis cs:14,307.815288689849)
--(axis cs:15,302.465093786994)
--(axis cs:16,301.745034847359)
--(axis cs:17,304.538296339734)
--(axis cs:18,301.295473146533)
--(axis cs:19,302.993745160217)
--(axis cs:19,277.806376060384)
--(axis cs:19,277.806376060384)
--(axis cs:18,277.442134936732)
--(axis cs:17,272.944411951654)
--(axis cs:16,278.588336885572)
--(axis cs:15,283.899654760092)
--(axis cs:14,286.553800244588)
--(axis cs:13,288.984081436589)
--(axis cs:12,282.942044872731)
--(axis cs:11,276.256685032187)
--(axis cs:10,279.401406676883)
--(axis cs:9,280.351803314573)
--(axis cs:8,282.072689189151)
--(axis cs:7,282.296647356635)
--(axis cs:6,280.679133409173)
--(axis cs:5,280.668215193962)
--(axis cs:4,280.493727369539)
--(axis cs:3,280.290696138718)
--(axis cs:2,279.461740021312)
--(axis cs:1,282.325895121208)
--(axis cs:0,288.110093240983)
--cycle;

\addplot [only marks, mark = +, draw=red, fill=red, colormap/viridis]
table{%
x                      y
0 302.34
1 294.8
2 290.57
3 289.45
4 290.37
5 294.11
6 286.43
7 293.5
8 293.56
9 288.67
10 291.95
13 302.37
14 299.18
15 289.95
16 291.67
18 288.85
19 290.68
};
\addplot [only marks, draw=red, fill=red, colormap/viridis]
table{%
x                      y
11 295.22
12 301.39
17 293.07
};
\addplot [semithick, color0, dashed]
table {%
11 262.592371879685
11 329.173911286602
};
\addplot [semithick, color0, dashed]
table {%
12 262.592371879685
12 329.173911286602
};
\addplot [semithick, color0, dashed]
table {%
17 262.592371879685
17 329.173911286602
};
\addplot [semithick, blue]
table {%
0 301.911715409259
1 294.842216927896
2 290.527031409001
3 290.087944887946
4 290.435229010133
5 290.717950204389
6 290.730726086886
7 292.281641845917
8 292.42767409138
9 290.609496945888
10 291.108566060677
11 293.417398001638
12 300.881958043682
13 302.201610368411
14 297.184544467218
15 293.182374273543
16 290.166685866466
17 288.741354145694
18 289.368804041633
19 290.4000606103
};
\end{axis}

\end{tikzpicture}
  \end{subfigure}  
  \hspace*{\fill}
  \begin{subfigure}[b]{.2\textwidth}
    \centering
\begin{tikzpicture}

\begin{axis}[
tick align=outside,
tick pos=left,
x grid style={white!69.01960784313725!black},
xmin=-0.5, xmax=19.5,
xtick style={color=black},
xlabel={View id},
y grid style={white!69.01960784313725!black},
ymin=-0.5, ymax=19.5,
ytick style={color=black},
y dir = reverse,
width=\figurewidth,
height=\figureheight
]
\addplot graphics [includegraphics cmd=\pgfimage,xmin=-0.5, xmax=19.5, ymin=19.5, ymax=-0.5] {./fig/tracker/002-K-000.png};
\end{axis}

\end{tikzpicture}  
  \end{subfigure}\\
  \begin{subfigure}[b]{.34\textwidth}
    \setlength{\figureheight}{\figurewidth}  
    \setlength{\figurewidth}{.93\textwidth}
    \centering
\begin{tikzpicture}

\definecolor{color0}{rgb}{0.12156862745098,0.466666666666667,0.705882352941177}

\begin{axis}[
legend cell align={left},
legend style={draw=white!80.0!black},
tick align=outside,
tick pos=left,
x grid style={white!69.01960784313725!black},
xlabel={View id},
xmin=-1.15600358422939, xmax=20.1560035842294,
xtick style={color=black},
y grid style={white!69.01960784313725!black},
ymin=324.918607212909, ymax=585.521637655098,
ytick style={color=black},
width=\figurewidth,
height=\figureheight
]
\path [draw=white!82.74509803921568!black, fill=white!82.74509803921568!black, opacity=0.5]
(axis cs:0,533.75731025809)
--(axis cs:0,573.676045362271)
--(axis cs:1,557.476195227764)
--(axis cs:2,490.780617933032)
--(axis cs:3,460.432236109955)
--(axis cs:4,431.474787051461)
--(axis cs:5,381.70038925662)
--(axis cs:6,373.187219398497)
--(axis cs:7,375.400644449304)
--(axis cs:8,384.160589014213)
--(axis cs:9,394.288216216744)
--(axis cs:10,403.447054651681)
--(axis cs:11,409.75355064217)
--(axis cs:12,415.009748777463)
--(axis cs:13,416.643641797198)
--(axis cs:14,421.430839984611)
--(axis cs:15,427.727602367648)
--(axis cs:16,426.325021704101)
--(axis cs:17,426.492434815869)
--(axis cs:18,430.789323505128)
--(axis cs:19,431.621570846255)
--(axis cs:19,390.634497168715)
--(axis cs:19,390.634497168715)
--(axis cs:18,391.511751608136)
--(axis cs:17,390.943020240517)
--(axis cs:16,389.623123090577)
--(axis cs:15,382.572614687039)
--(axis cs:14,384.566595071237)
--(axis cs:13,384.26787592723)
--(axis cs:12,381.129917476463)
--(axis cs:11,375.907367627266)
--(axis cs:10,371.372520251598)
--(axis cs:9,362.983976757844)
--(axis cs:8,352.10759655471)
--(axis cs:7,341.550500566409)
--(axis cs:6,336.764199505736)
--(axis cs:5,343.864916797038)
--(axis cs:4,396.673633542806)
--(axis cs:3,421.908310765311)
--(axis cs:2,434.487856994936)
--(axis cs:1,517.663988203896)
--(axis cs:0,533.75731025809)
--cycle;

\path [fill=color0, fill opacity=0.5]
(axis cs:0,541.082916720654)
--(axis cs:0,566.350438899707)
--(axis cs:1,550.1505887652)
--(axis cs:2,483.455011470468)
--(axis cs:3,453.106629647392)
--(axis cs:4,424.149180588898)
--(axis cs:5,374.374782794057)
--(axis cs:6,365.861612935933)
--(axis cs:7,368.07503798674)
--(axis cs:8,376.83498255165)
--(axis cs:9,386.96260975418)
--(axis cs:10,396.121448189117)
--(axis cs:11,402.427944179606)
--(axis cs:12,407.684142314899)
--(axis cs:13,409.318035334634)
--(axis cs:14,414.105233522047)
--(axis cs:15,420.401995905084)
--(axis cs:16,418.999415241537)
--(axis cs:17,419.166828353305)
--(axis cs:18,423.463717042565)
--(axis cs:19,424.295964383691)
--(axis cs:19,397.960103631279)
--(axis cs:19,397.960103631279)
--(axis cs:18,398.8373580707)
--(axis cs:17,398.268626703081)
--(axis cs:16,396.948729553141)
--(axis cs:15,389.898221149603)
--(axis cs:14,391.8922015338)
--(axis cs:13,391.593482389793)
--(axis cs:12,388.455523939027)
--(axis cs:11,383.23297408983)
--(axis cs:10,378.698126714162)
--(axis cs:9,370.309583220408)
--(axis cs:8,359.433203017274)
--(axis cs:7,348.876107028973)
--(axis cs:6,344.0898059683)
--(axis cs:5,351.190523259602)
--(axis cs:4,403.99924000537)
--(axis cs:3,429.233917227874)
--(axis cs:2,441.8134634575)
--(axis cs:1,524.98959466646)
--(axis cs:0,541.082916720654)
--cycle;

\addplot [only marks, mark = +, draw=red, fill=red, colormap/viridis]
table{%
x                      y
0 564.51
1 531.84
3 444.71
4 416.7
5 351.97
7 367.57
8 366.69
9 371.76
10 390.11
11 395.4
12 395.63
13 398.78
14 405.56
16 406.62
17 409.37
18 411.51
19 410.75
};
\addplot [only marks, draw=red, fill=red, colormap/viridis]
table{%
x                      y
2 491.14
6 361.75
15 408.77
};
\addplot [semithick, color0, dashed, forget plot]
table {%
2 324.918607212909
2 585.521637655098
};
\addplot [semithick, color0, dashed, forget plot]
table {%
6 324.918607212909
6 585.521637655098
};
\addplot [semithick, color0, dashed, forget plot]
table {%
15 324.918607212909
15 585.521637655098
};
\addplot [semithick, blue]
table {%
0 553.716677810181
1 537.57009171583
2 462.634237463984
3 441.170273437633
4 414.074210297134
5 362.782653026829
6 354.975709452116
7 358.475572507856
8 368.134092784462
9 378.636096487294
10 387.40978745164
11 392.830459134718
12 398.069833126963
13 400.455758862214
14 402.998717527924
15 405.150108527343
16 407.974072397339
17 408.717727528193
18 411.150537556632
19 411.128034007485
};
\end{axis}

\end{tikzpicture}
    \caption{GP regression of \\ $u$ coordinates}
  \end{subfigure}
  \hspace*{\fill}
  \begin{subfigure}[b]{.34\textwidth}
    \setlength{\figureheight}{\figurewidth}  
    \setlength{\figurewidth}{.93\textwidth}
    \centering
\begin{tikzpicture}

\definecolor{color0}{rgb}{0.12156862745098,0.466666666666667,0.705882352941177}

\begin{axis}[
tick align=outside,
tick pos=left,
x grid style={white!69.01960784313725!black},
xlabel={View id},
xmin=-1.15600358422939, xmax=20.1560035842294,
xtick style={color=black},
y grid style={white!69.01960784313725!black},
ymin=171.502395973232, ymax=253.141559573495,
ytick style={color=black},
width=\figurewidth,
height=\figureheight
]
\path [draw=white!82.74509803921568!black, fill=white!82.74509803921568!black, opacity=0.5]
(axis cs:0,175.213267045972)
--(axis cs:0,215.132002150152)
--(axis cs:1,218.546371908457)
--(axis cs:2,245.513341236254)
--(axis cs:3,238.045526723068)
--(axis cs:4,238.044354113649)
--(axis cs:5,245.375132947017)
--(axis cs:6,247.111709486913)
--(axis cs:7,246.629744005433)
--(axis cs:8,244.520767887916)
--(axis cs:9,241.547295211706)
--(axis cs:10,239.138758346743)
--(axis cs:11,238.126660623268)
--(axis cs:12,236.700412324738)
--(axis cs:13,235.207546874618)
--(axis cs:14,236.671313914647)
--(axis cs:15,240.47491959286)
--(axis cs:16,240.739527377348)
--(axis cs:17,242.658293520939)
--(axis cs:18,249.209397443368)
--(axis cs:19,249.430688500756)
--(axis cs:19,208.443614823217)
--(axis cs:19,208.443614823217)
--(axis cs:18,209.931825546375)
--(axis cs:17,207.108878945588)
--(axis cs:16,204.037628763825)
--(axis cs:15,195.319931912251)
--(axis cs:14,199.807069001272)
--(axis cs:13,202.83178100465)
--(axis cs:12,202.820581023739)
--(axis cs:11,204.280477608364)
--(axis cs:10,207.06422394666)
--(axis cs:9,210.243055752807)
--(axis cs:8,212.467775428413)
--(axis cs:7,212.779600122538)
--(axis cs:6,210.688689594152)
--(axis cs:5,207.539660487435)
--(axis cs:4,203.243200604994)
--(axis cs:3,199.521601378424)
--(axis cs:2,189.220580298158)
--(axis cs:1,178.73416488459)
--(axis cs:0,175.213267045972)
--cycle;

\path [fill=color0, fill opacity=0.5]
(axis cs:0,182.538873508535)
--(axis cs:0,207.806395687588)
--(axis cs:1,211.220765445893)
--(axis cs:2,238.18773477369)
--(axis cs:3,230.719920260505)
--(axis cs:4,230.718747651085)
--(axis cs:5,238.049526484453)
--(axis cs:6,239.786103024349)
--(axis cs:7,239.304137542869)
--(axis cs:8,237.195161425352)
--(axis cs:9,234.221688749142)
--(axis cs:10,231.813151884179)
--(axis cs:11,230.801054160704)
--(axis cs:12,229.374805862174)
--(axis cs:13,227.881940412054)
--(axis cs:14,229.345707452083)
--(axis cs:15,233.149313130297)
--(axis cs:16,233.413920914784)
--(axis cs:17,235.332687058376)
--(axis cs:18,241.883790980804)
--(axis cs:19,242.105082038192)
--(axis cs:19,215.76922128578)
--(axis cs:19,215.76922128578)
--(axis cs:18,217.257432008939)
--(axis cs:17,214.434485408152)
--(axis cs:16,211.363235226389)
--(axis cs:15,202.645538374815)
--(axis cs:14,207.132675463836)
--(axis cs:13,210.157387467214)
--(axis cs:12,210.146187486302)
--(axis cs:11,211.606084070928)
--(axis cs:10,214.389830409224)
--(axis cs:9,217.56866221537)
--(axis cs:8,219.793381890977)
--(axis cs:7,220.105206585101)
--(axis cs:6,218.014296056716)
--(axis cs:5,214.865266949999)
--(axis cs:4,210.568807067557)
--(axis cs:3,206.847207840988)
--(axis cs:2,196.546186760722)
--(axis cs:1,186.059771347153)
--(axis cs:0,182.538873508535)
--cycle;

\addplot [only marks, mark = +, draw=red, fill=red, colormap/viridis]
table{%
x                      y
0 193.32
1 199.53
3 218
4 221.91
5 224.93
7 230.89
8 230.6
9 223.35
10 223.76
11 220.09
12 221.79
13 218.36
14 217.55
16 222.38
17 225.11
18 229.66
19 229.13
};
\label{addplot:marker-plus}
\addplot [only marks, draw=red, fill=red, colormap/viridis]
table{%
x                      y
2 214.02
6 229.66
15 219.94
};
\label{addplot:marker-bullet}
\addplot [semithick, color0, dashed]
table {%
2 171.502395973232
2 253.141559573495
};
\addplot [semithick, color0, dashed]
table {%
6 171.502395973232
6 253.141559573495
};
\addplot [semithick, color0, dashed]
table {%
15 171.502395973232
15 253.141559573495
};
\addplot [semithick, blue]
table {%
0 195.172634598062
1 198.640268396523
2 217.366960767206
3 218.783564050746
4 220.643777359321
5 226.457396717226
6 228.900199540532
7 229.704672063985
8 228.494271658164
9 225.895175482256
10 223.101491146701
11 221.203569115816
12 219.760496674238
13 219.019663939634
14 218.239191457959
15 217.897425752556
16 222.388578070586
17 224.883586233264
18 229.570611494871
19 228.937151661986
};
\end{axis}

\end{tikzpicture}
    \caption{GP regression of \\ $v$ coordinates}
  \end{subfigure}  
  \hspace*{\fill}
  \begin{subfigure}[b]{.2\textwidth}
    \centering
\begin{tikzpicture}

\begin{axis}[
tick align=outside,
tick pos=left,
x grid style={white!69.01960784313725!black},
xmin=-0.5, xmax=19.5,
xtick style={color=black},
xlabel={View id},
y grid style={white!69.01960784313725!black},
ymin=-0.5, ymax=19.5,
ytick style={color=black},
y dir = reverse,
width=\figurewidth,
height=\figureheight
]
\addplot graphics [includegraphics cmd=\pgfimage,xmin=-0.5, xmax=19.5, ymin=19.5, ymax=-0.5] {./fig/tracker/003-K-000.png};
\end{axis}

\end{tikzpicture}  
   \caption{Pose \\ covariance }
  \end{subfigure}

  \caption{GP regression results of three tracks (out of 533) using the pose kernel in  \cref{sec:track}. The red points corresponds to ground-truth trajectories, where `\ref{addplot:marker-plus}' means training points and `\ref{addplot:marker-bullet}' corresponds to unseen points. The blue points are predicted mean values. The shaded patches denote the 95\% quantiles.}
  \label{fig:tracres}
\end{figure*}
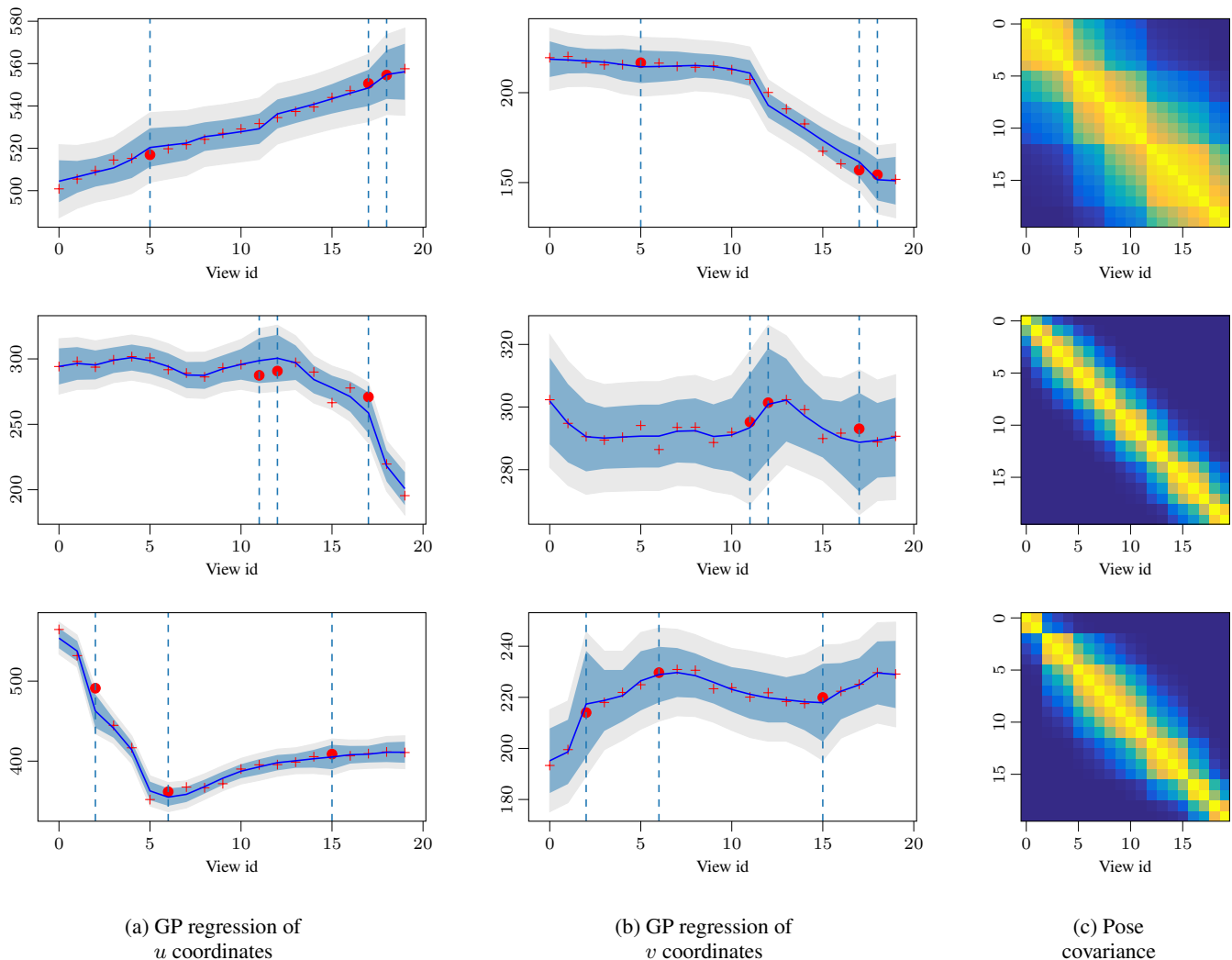

Before regression, for the training points in each track, we subtract the mean of the track so that each track will have zero mean, and we add the mean back after regression. \cref{fig:tracres} shows GP regression results of three tracks and the prior pose covariance in each respective case. For each trajectory, there are three randomly chosen missing points, and we predict the $u$ and $v$ pixel coordinates for the missing points separately by using the same pose kernel in the GP prior. The red points corresponds to original data points, where `\ref{addplot:marker-plus}' indicates training points and `\ref{addplot:marker-bullet}' corresponds to unseen test points. The blue line is the GP posterior mean.

\section{Details on the View Synthesis Experiment}
\label{app:gppvae}
In the experiments in \cref{sec:gppvae}, we used a similar setup as in GPPVAE with joint optimization (GPPVAE-joint) in \citet{Casale+Dalca+Saglietti+Listgarten+Fusi:2018}, where one first trains the standard VAE and then optimizes the autoencoder and GP parameters jointly.  For the VAE model, the dimension of the latent space is $L =256$ and the convolutional architecture is the same as in \citet{Casale+Dalca+Saglietti+Listgarten+Fusi:2018}. For the object kernel, we set the dimensionality of the object feature vectors to $M = 128$. For the view kernel, we follow the \cref{eq:viewx}, where the lengthscale hyperparameters $\ell_\mathrm{x}, \ell_\mathrm{y}, \ell_\mathrm{z}$ in the diagonal matrix $\MLambda$ are learned during training. The whole model is implemented in PyTorch. The standard VAE was trained with 1000 epochs and the GP parameters were trained jointly for 200 epochs with the batch size of 64.

\begin{figure*}[!t]
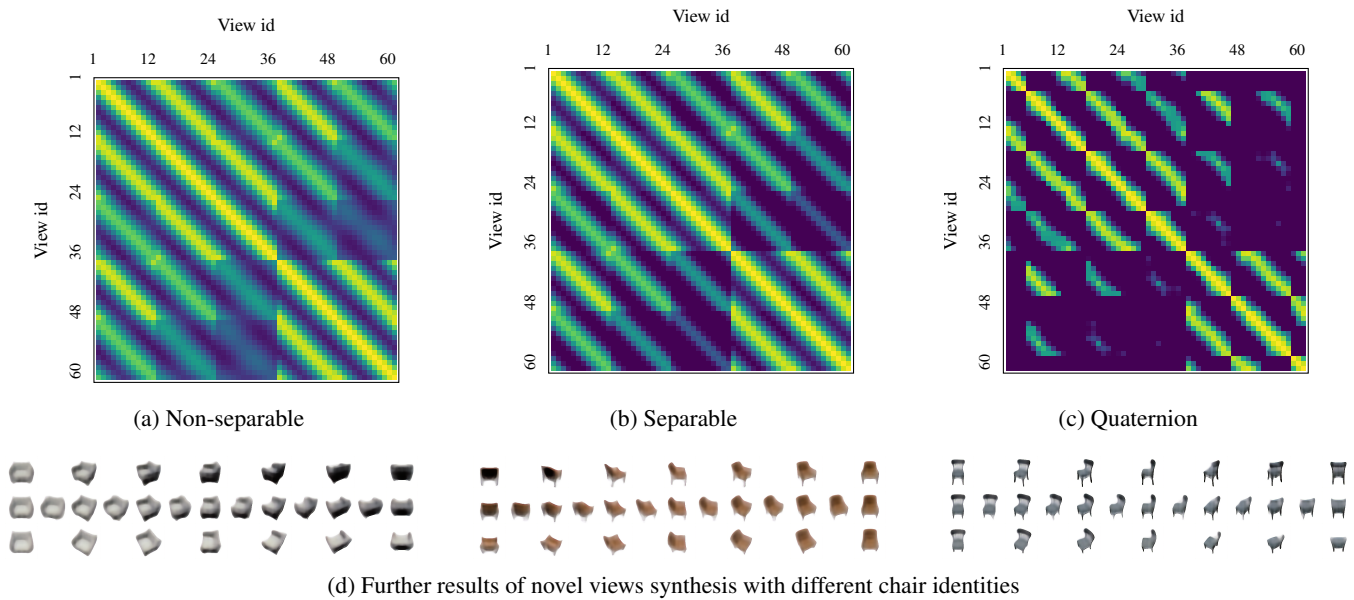

  \centering\scriptsize
  \pgfplotsset{yticklabel style={rotate=90}, ylabel style={yshift=-15pt},clip=true,scale only axis,axis on top,clip marker paths,legend style={row sep=0pt},legend columns=-1,xlabel near ticks,clip=false}
  \setlength{\figurewidth}{.2\textwidth}
  \setlength{\figureheight}{\figurewidth}
  \begin{subfigure}[b]{.32\textwidth} 
    \centering 
    \begin{tikzpicture}[inner sep=0]
       \foreach \i in {1,12,24,36, 48, 60} {
         \node[] at ({-0.35*\textwidth+\i/60*0.7*\textwidth-0.7*\textwidth/36},0.4\textwidth) {\tiny \i};
         \node[rotate=90] at (-0.4\textwidth,{0.35*\textwidth-\i/60*0.7*\textwidth+0.7*\textwidth/36}) {\tiny \i};
       }
       \node[] at (0,0.48\textwidth) {View id};
       \node[rotate=90] at (-0.48\textwidth,0) {View id};       
       \node[draw=black,inner sep=1pt] at (0,0) {\includegraphics[width=.7\textwidth]{fig/chairs/supp/nonsep_cov}};
    \end{tikzpicture} 
    \vspace*{.5em}
    \caption{Non-separable}  
    \label{fig:chair-nonsep-cov}    
  \end{subfigure}
  \hspace*{\fill}
  \begin{subfigure}[b]{.32\textwidth} 
    \centering
    \begin{tikzpicture}[inner sep=0]
       \foreach \i in {1,12,24,36, 48, 60} {
         \node[] at ({-0.35*\textwidth+\i/60*0.7*\textwidth-0.7*\textwidth/36},0.4\textwidth) {\tiny \i};
         \node[rotate=90] at (-0.4\textwidth,{0.35*\textwidth-\i/60*0.7*\textwidth+0.7*\textwidth/36}) {\tiny \i};
       }
       \node[] at (0,0.48\textwidth) {View id};
       \node[rotate=90] at (-0.48\textwidth,0) {View id};       
       \node[draw=black,inner sep=1pt] at (0,0) {\includegraphics[width=.7\textwidth]{fig/chairs/supp/sep_cov}};
    \end{tikzpicture}
    \vspace*{1em}
    \caption{Separable}  
    \label{fig:chair-sep-cov}    
  \end{subfigure}
  \hspace*{\fill}  
  \begin{subfigure}[b]{.32\textwidth} 
    \centering 

    \begin{tikzpicture}[inner sep=0]
       \foreach \i in {1,12,24,36, 48, 60} {
         \node[] at ({-0.35*\textwidth+\i/60*0.7*\textwidth-0.7*\textwidth/36},0.4\textwidth) {\tiny \i};
         \node[rotate=90] at (-0.4\textwidth,{0.35*\textwidth-\i/60*0.7*\textwidth+0.7*\textwidth/36}) {\tiny \i};
       }
       \node[] at (0,0.48\textwidth) {View id};
       \node[rotate=90] at (-0.48\textwidth,0) {View id};       
       \node[draw=black,inner sep=1pt] at (0,0) {\includegraphics[width=.7\textwidth]{fig/chairs/supp/quat_cov}};
    \end{tikzpicture}
    \vspace*{1em}
    \caption{Quaternion}  
    \label{fig:chair-quat-cov}
  \end{subfigure}  
  \\[1em]
  \begin{subfigure}{\textwidth} 
  \begin{minipage}[t]{.3\textwidth}      
    \tiny

    \setlength{\figurewidth}{.079\textwidth}
    \setlength{\figureheight}{.9\figurewidth}
    
      \begin{tikzpicture}[inner sep=0]
      \foreach \x [count=\i] in {0012,0013,0014, 0015,0016,0017, 0018} 
      {

        \node[draw=white,fill=black!20,minimum size=\figurewidth,inner sep=0pt]
        (\i) at ({2*\figurewidth*\i},{-\figureheight*1})
        {\includegraphics[width=\figurewidth]{./fig/chairs/supp/oos/116/reco/\x}};
      }
    \end{tikzpicture}\\
    \begin{tikzpicture}[inner sep=0]
      \foreach \x [count=\i] in {0000,0002,0005, 0007,0010,0012,0015,0017,0020,
0022,0025,0027, 0030} 
      {

        \node[draw=white,fill=black!20,minimum size=\figurewidth,inner sep=0pt]
        (\i) at ({\figurewidth*\i},{-\figureheight*1})
        {\includegraphics[width=\figurewidth]{./fig/chairs/supp/oos/116/pred/\x}};
      }
    \end{tikzpicture}\\
    \begin{tikzpicture}[inner sep=0]
      \foreach \x [count=\i] in {0024,0025,0026, 0027,0028,0029, 0030} 
      {

        \node[draw=white,fill=black!20,minimum size=\figurewidth,inner sep=0pt]
        (\i) at ({2*\figurewidth*\i},{-\figureheight*1})
        {\includegraphics[width=\figurewidth]{./fig/chairs/supp/oos/116/reco/\x}};
      }
    \end{tikzpicture}
  \end{minipage}\hfill
  \begin{minipage}[t]{.3\textwidth}      
    \tiny

    \setlength{\figurewidth}{.079\textwidth}
    \setlength{\figureheight}{.9\figurewidth}      
    \begin{tikzpicture}[inner sep=0]
      \foreach \x [count=\i] in {0018,0019,0020, 0021,0022,0023, 0012} 
      {

        \node[draw=white,fill=black!20,minimum size=\figurewidth,inner sep=0pt]
        (\i) at ({2*\figurewidth*\i},{-\figureheight*1})
        {\includegraphics[width=\figurewidth]{./fig/chairs/supp/oos/48/reco/\x}};
      }
    \end{tikzpicture}\\
    \begin{tikzpicture}[inner sep=0]
      \foreach \x [count=\i] in {0030,0032,0035, 0037,0040,0042,0045,0047,0050,
0052,0055,0057, 0000} 
      {

        \node[draw=white,fill=black!20,minimum size=\figurewidth,inner sep=0pt]
        (\i) at ({\figurewidth*\i},{-\figureheight*1})
        {\includegraphics[width=\figurewidth]{./fig/chairs/supp/oos/48/pred/\x}};
      }
    \end{tikzpicture}\\
    \begin{tikzpicture}[inner sep=0]
      \foreach \x [count=\i] in {0030,0031,0032, 0033,0034,0035, 0024} 
      {

        \node[draw=white,fill=black!20,minimum size=\figurewidth,inner sep=0pt]
        (\i) at ({2*\figurewidth*\i},{-\figureheight*1})
        {\includegraphics[width=\figurewidth]{./fig/chairs/supp/oos/48/reco/\x}};
      }
    \end{tikzpicture}
  \end{minipage}\hfill
  \begin{minipage}[t]{.3\textwidth}      
    \tiny

    \setlength{\figurewidth}{.079\textwidth}
    \setlength{\figureheight}{.9\figurewidth}           
    \begin{tikzpicture}[inner sep=0]
      \foreach \x [count=\i] in {0012,0013,0014, 0015,0016,0017, 0018} 
      {

        \node[draw=white,fill=black!20,minimum size=\figurewidth,inner sep=0pt]
        (\i) at ({2*\figurewidth*\i},{-\figureheight*1})
        {\includegraphics[width=\figurewidth]{./fig/chairs/supp/oos/125/reco/\x}};
      }
    \end{tikzpicture}\\
    \begin{tikzpicture}[inner sep=0]
      \foreach \x [count=\i] in {0000,0002,0005, 0007,0010,0012,0015,0017,0020,
0022,0025,0027, 0030} 
      {

        \node[draw=white,fill=black!20,minimum size=\figurewidth,inner sep=0pt]
        (\i) at ({\figurewidth*\i},{-\figureheight*1})
        {\includegraphics[width=\figurewidth]{./fig/chairs/supp/oos/125/pred/\x}};
      }
    \end{tikzpicture}\\
    \begin{tikzpicture}[inner sep=0]
      \foreach \x [count=\i] in {0024,0025,0026, 0027,0028,0029, 0030} 
      {

        \node[draw=white,fill=black!20,minimum size=\figurewidth,inner sep=0pt]
        (\i) at ({2*\figurewidth*\i},{-\figureheight*1})
        {\includegraphics[width=\figurewidth]{./fig/chairs/supp/oos/125/reco/\x}};
      }
    \end{tikzpicture}
    \end{minipage}    
    \caption{Further results of novel views synthesis with different chair identities}
    \label{fig:chairs-supp}    
  \end{subfigure}

  \caption{Results from experiments on the ShapeNet chairs data set. (a)~Non-separable covariance matrix computed with rotation matrices for 60 training views. (b)~Separable covariance matrix computed with Euler angles for 60 training views. (c)~Quaternion covariance matrix for 60 training views. (d)~Results of novel views synthesis with different chairs. For each chair, the first (elevation $30^{\circ}$) and the third row (elevation $60^{\circ}$) show predictions for views from angles found in the training set, while the whole second row shows predictions for angles not in the training set (elevation $45^{\circ}$). The limited representation power of the model is due to the GPPVAE framework.}
  \label{fig:more-chairs}
\end{figure*}

\cref{fig:more-chairs} provides more results on the ShapeNet chair data set. \cref{fig:chair-nonsep-cov}, \cref{fig:chair-sep-cov} and \cref{fig:chair-quat-cov} show the difference between the non-separable kernel, the separable kernel and the quaternion kernel we discussed in \cref{sec:view-kernels}. Generally, non-separable kernel and separable kernel show similar patterns and lead to comparable quantitive results, while for widely separated views, such as elevation $60^{\circ}$ and elevation $-60^{\circ}$, the separable kernel has weaker covariance than the non-separable kernel. Because of the non-uniqueness of quaternions, the quaternion kernel in \cref{fig:chair-quat-cov} shows different patterns and leads to worse quantitive results. Given different chair IDs, \cref{fig:chairs-supp} presents novel view prediction results with different kinds of chairs.

\section{Details on the Face Reconstruction Experiment}
\label{app:face}
For the experiments in \cref{sec:face}, we started from recording a single MPEG movie file for each face identity, 20--30 seconds each, with associated camera poses (from Apple ARKit) captured on an iPhone XS. We decomposed the movie into a sequence of single image frames that we crop to $1024 {\times} 1024$ and aligned the detected faces therein using the approach of \citet{Karras+Laine+Aila:2019} based on \citet{kazemi:2015}. For each image, we then created the corresponding StyleGAN~\cite{Karras+Laine+Aila:2019} latent representation using a script based on \citet{puzer:2019}, with learning rate 0.01 and 200 iterations per image, as follows. We first generated a corresponding initial `guess' values for the $18 {\times} 512$ latent variable matrix that StyleGAN generator uses to produce a $1024 {\times} 1024$ image. We fed both the generated image and the real camera image to a pre-trained VGG-16~\cite{simonyan:2015} network. We read off the VGG-16 feature space representations of both images, and used the difference as a loss to drive an optimization process that, upon convergence, produces the latent variable value that can be used to reproduce an image that closely resembles the original camera image. The StyleGAN model was pre-trained on the FFHQ dataset of \citet{Karras+Laine+Aila:2019}. The projections were implemented in Tensorflow while the GP part was implemented in PyTorch.

This rather heuristic projection process is very slow (60--120 seconds per image on a Titan~V GPU), but usually produces high quality images (while other generative models with built-in encoders could have produced results 50--100 times faster, but typically lower quality) with only occasional visible artifacts. The weaknesses of the model can be seen when failing to properly reconstruct face shots where the azimuth rotation angle is large. We run the reconstruction process for every fifth frame of the original video. Some small segments of some of the videos were excluded from beginning or end when the reconstruction failed completely.

We then carried out three experiments. First, we used the latent codes of all reconstructed images (total of 50--200 frames, depending on original video length and face identity) to construct a single matrix comprising all the training samples. We then ran the GP regression on the matrix in the usual manner using our view-aware prior, and the camera pose data from the original video frames. This resulted in a reconstructed latent variable matrix of the same size as before. We then fed these latent variables back to StyleGAN and decoded each code back to the image space, resulting in smoothed versions of each of the original frames. The smoothing effect can be seen in the attached video.

\begin{figure*}[!t]
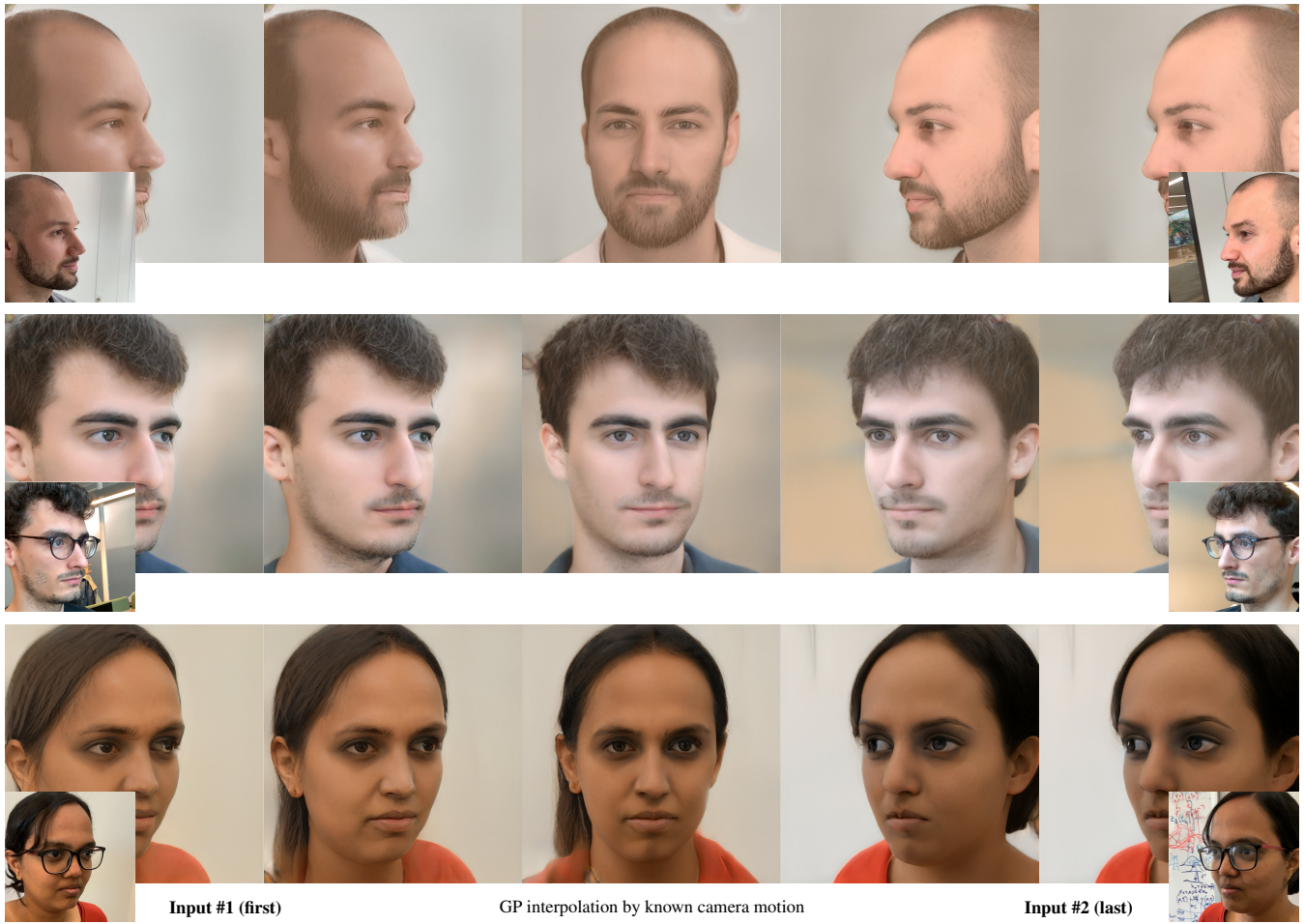

  \centering\scriptsize
  \pgfplotsset{yticklabel style={rotate=90}, ylabel style={yshift=-15pt},clip=true,scale only axis,axis on top,clip marker paths,legend style={row sep=0pt},legend columns=-1,xlabel near ticks}
  \setlength{\figurewidth}{.2\textwidth}
  \setlength{\figureheight}{\figurewidth}
  \begin{tikzpicture}[inner sep=0]
    \foreach \x [count=\i] in {
      stylegan_interpolations/0406,    
      stylegan_interpolations/0301,
      stylegan_interpolations/0201,
      stylegan_interpolations/0101,
      stylegan_interpolations/0001} 
      {
        \node[draw=white,fill=black!20,minimum size=\figurewidth,inner sep=0pt]
        (\i) at ({\figurewidth*\i},{-\figureheight*0})
        {\includegraphics[width=\figurewidth]{fig/video-interpolation/face05/\x.jpg}};
      }
     \foreach \x [count=\i] in {stylegan_corner_reconstruction/1076,
      stylegan_interpolations/0251,
      stylegan_interpolations/0151,
      stylegan_interpolations/0051,
      stylegan_corner_reconstruction/0726} 
      {
        \node[draw=white,fill=black!20,minimum size=\figurewidth,inner sep=0pt]
        (\i) at ({\figurewidth*\i},{-\figureheight*2.4})
        {\includegraphics[width=\figurewidth]{fig/video-interpolation/face03/\x.jpg}};
      }
    \foreach \x [count=\i] in {
      stylegan_interpolations/0371,    
      stylegan_interpolations/0271,
      stylegan_interpolations/0181,
      stylegan_interpolations/0091,
      stylegan_interpolations/0001} 
      {
        \node[draw=white,fill=black!20,minimum size=\figurewidth,inner sep=0pt]
        (\i) at ({\figurewidth*\i},{-\figureheight*1.2})
        {\includegraphics[width=\figurewidth]{fig/video-interpolation/face02/\x.jpg}};
      }

    \node[draw=white,fill=black!20,minimum size=.5\figurewidth,inner sep=0pt]
        (i) at ({0.75*\figureheight},{-0.4*\figureheight})
        {\includegraphics[width=.5\figurewidth]{fig/video-interpolation/face05/corner_aligned/0806.jpg}};
    \node[draw=white,fill=black!20,minimum size=.5\figurewidth,inner sep=0pt]
        (i) at ({5.25*\figureheight},{-0.4*\figureheight})
        {\includegraphics[width=.5\figurewidth]{fig/video-interpolation/face05/corner_aligned/0401.jpg}};

    \node[draw=white,fill=black!20,minimum size=.5\figurewidth,inner sep=0pt]
        (i) at ({0.75*\figureheight},{-2.8*\figureheight})
        {\includegraphics[width=.5\figurewidth]{fig/video-interpolation/face03/corner_aligned/1076.jpg}};
    \node[draw=white,fill=black!20,minimum size=.5\figurewidth,inner sep=0pt]
        (i) at ({5.25*\figureheight},{-2.8*\figureheight})
        {\includegraphics[width=.5\figurewidth]{fig/video-interpolation/face03/corner_aligned/0726.jpg}};

    \node[draw=white,fill=black!20,minimum size=.5\figurewidth,inner sep=0pt]
        (i) at ({0.75*\figureheight},{-1.6*\figureheight})
        {\includegraphics[width=.5\figurewidth]{fig/video-interpolation/face02/corner_aligned/0976.jpg}};
    \node[draw=white,fill=black!20,minimum size=.5\figurewidth,inner sep=0pt]
        (i) at ({5.25*\figureheight},{-1.6*\figureheight})
        {\includegraphics[width=.5\figurewidth]{fig/video-interpolation/face02/corner_aligned/0606.jpg}};

    \node at (1.35\figureheight,-3.0\figureheight) {\bf Input \#1 (first)};
    \node at (4.65\figureheight,-3.0\figureheight) {\bf Input \#2 (last)};
    \node at (3\figureheight,-3.0\figureheight) {GP interpolation by known camera motion};
        
  \end{tikzpicture}
  \caption{Three more examples of view-aware latent space interpolation using the latent space of StyleGAN. These reconstructions are based on only taking the first and the last frame of short side-to-side video sequences (Input \#1 and \#2), encoding them into the GAN latent space, and interpolating the intermediate frames using only the information of the associated camera poses (Apple ARKit on an iPhone~XS). The intermediate frames were recovered by regressing the latent space with our view-aware GP prior. The frames are re-created in correct head orientations. The irregular angular speed of the camera movement is precisely captured by our method, resulting in non-symmetric interpolation.}
  \label{fig:interpolation-all}
  \vspace*{-1em}
\end{figure*}

Second, we set up an essentially similar experiment, but now using the latent codes of merely the start and the end frame (\ie, two frames only, resulting in $2 {\times} 18 {\times} 512$ latent matrix) of the sequence, in combination with the full covariance matrix of the camera pose. Nonetheless, we used the same method as above to reproduce the whole sequence of latent codes (\eg, $50 {\times} 18 {\times} 512$). That is, all the latent variables between the start and end frames were interpolated with the view-aware GP prior. Again, we fed the resulting latent codes back to StyleGAN, and confirmed that the resulting frames not only have high quality, but also precisely follow the camera movement (see \cref{fig:interpolation-all}).

Finally, to evaluate the specific contribution of the view-aware GP prior, we tried a simpler experiment where the same start and end frames were used as an input, but the intermediate latent codes were produced by simple linear interpolation in the latent space (\ie, just taking latent values from evenly sized intervals between the two known latent values). We again fed these to StyleGAN to produce the corresponding images. As should be expected (if the latent space is well-behaved), the resulting images still have high quality, but they simply rotate the face in evenly sized increments per frame. This should be contrasted with the previous experiment with view-aware prior, in which the frames actually match the original non-linear camera movement. For comparison, see \cref{fig:interpolation_comp}.

The LPIPS results on table \cref{tbl:lpips} were created as follows. For the four camera runs on different face identities, we selected the same start and end points as in the experiments above, so as to have roughly symmetric start and end frames. This setup facilitates comparison to linear interpolation. We used every 5th frame as before, presuming that consecutive ground-truth frames are nearly identical, with 1570 total actually evaluated frames. We cropped each frame around the center as in the LPIPS experiments of \cite{Karras+Laine+Aila:2019} and compute the LPIPS between the reconstructed (and smoothed or interpolated) frames and the corresponding original video frames. When all frames are used, StyleGAN projections are simply reconstructing each frame independently, and one expects this mode to yield best LPIPS. GP smoothing also uses all frames, but smoothens the differences between consecutive frames. Finally, linear interpolation and GP interpolation are used by leveraging {\textit{only}} the first and the last frame; however, the GP approach by definition also uses the kernel. Hence, only the four bottom rows are directly comparable. We emphasize that these measures are very dependent on the exact way we set up the experiment. Finally, for each setup, we measure the smoothness of the changes of the sequence itself by measuring the LPIPS between each frame at $t$ and its follow-up frame at $t+1$ within the same sequence, and take the mean. The smaller the average change, the less jitter there is in the video that corresponds to that sequence of frames. This measure (LPIPS-$\Delta$) of course only relevant if the LPIPS distance between each individual interpolated frame and the original (or directly GAN projected) frame is low, since otherwise one could minimize LPIPS-$\Delta$ by simply making each consequtive frame identical. Note that our goal is to show that the GP interpolation approach is working as expected, not to claim that it is the best method for this specific task.

\begin{figure*}[!t]
  \centering\scriptsize
  \setlength{\figurewidth}{.14\textwidth}
  \setlength{\figureheight}{\figurewidth}
  \resizebox{\textwidth}{!}{%
  \begin{tikzpicture}[inner sep=0]

    \foreach \y [count=\j] in {fig/video-interpolation/face02/original,fig/video-interpolation/face02/reconstructed_aligned,fig/video-interpolation/face02/stylegan_interpolations, fig/video-interpolation/face02/linear, fig/video-interpolation/face02/euler_interpolations, fig/video-interpolation/face02/quat_interpolations} {

      \foreach \x [count=\i] in {0001,0046,0091,0136,0181,0226,0271,0316,0371} {
          \node[draw=white,fill=black!20,minimum size=\figurewidth,inner sep=0pt]
          (\i) at ({-\figurewidth*\i},{-1*\figureheight*\j})
          {\includegraphics[width=\figurewidth]{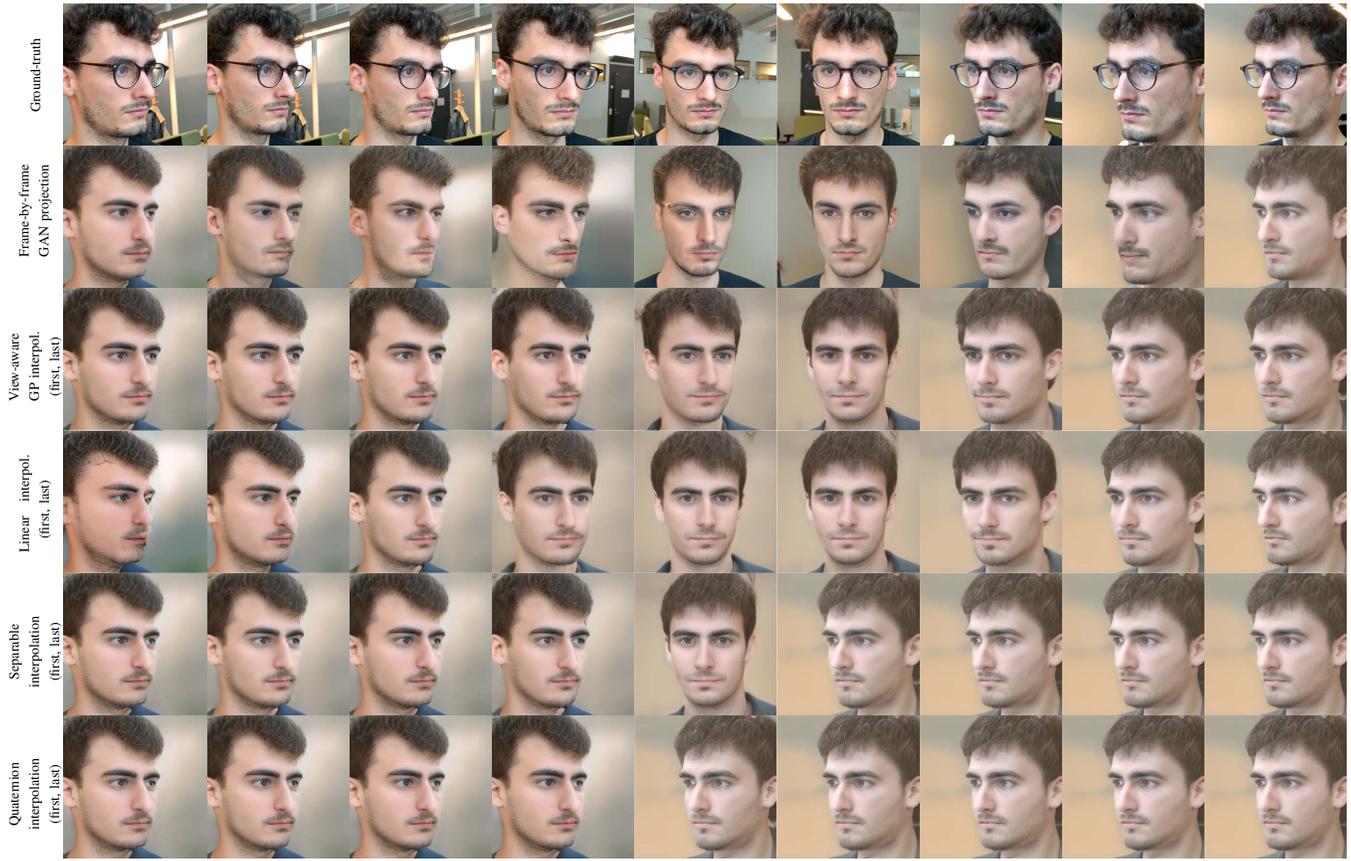}};
      }
    }

    \node[rotate=90,text width=0.8*\figureheight,align=center] at (-9.7*\figurewidth,-1*\figureheight) {\scriptsize Ground-truth};
    \node[rotate=90,text width=0.9*\figureheight,align=center] at (-9.7*\figurewidth,-1.95*\figureheight) {\scriptsize Frame-by-frame GAN projection};

    \node[rotate=90,text width=0.9*\figureheight,align=center] at (-9.7*\figurewidth,-3.05*\figureheight) {\scriptsize View-aware \mbox{GP interpol.} \mbox{(first, last)}};
    \node[rotate=90,text width=0.8*\figureheight,align=center] at (-9.7*\figurewidth,-4.05*\figureheight) {\scriptsize \mbox{~~Linear~~} \mbox{interpol.} \mbox{(first, last)}};
    \node[rotate=90,text width=0.9*\figureheight,align=center] at (-9.7*\figurewidth,-5.05*\figureheight) {\scriptsize Separable \mbox{interpolation} \mbox{(first, last)}};
    \node[rotate=90,text width=0.9*\figureheight,align=center] at (-9.7*\figurewidth,-6.05*\figureheight) {\scriptsize Quaternion \mbox{interpolation} \mbox{(first, last)}};
  \end{tikzpicture}}

  \caption{View-aware GP interpolation between two input frames, including comparisons to separable kernel and Quaternion kernel: {\bf Row~\#1:}~Frames separated by equal time intervals from a camera run, aligned on the face. {\bf Row~\#2:}~Each frame independently reconstructed by the GAN. {\bf Row~\#3:}~Interpolation in GAN latent space between first and last frame by our view-aware GP prior. {\bf Row~\#4:}~Linear interpolation of the intermediate frames in GAN latent space between first and last frame (note the lost azimuth angle). {\bf Row~\#5:}~Interpolation by the separable baseline kernel.~{\bf Row~\#6:}~Interpolation by the Quaternion baseline kernel. The linear interpolation behaves artifiallly well here since the original sequence happens to be close to linear; however, the error in the middle frame highlights the subtle departure from non-linearity in the original frames, captured perfectly by our view-aware kernel, but obviously missed by the linear interpolation.}
  \label{fig:interpolation_comp_euler_int}
  \vspace*{-1em}
\end{figure*}

Comparisons to baseline kernels using the separable kernel and quaternion kernel are visually illustrated in \cref{fig:interpolation_comp_euler_int}, clearly showing the jerky and out-of-phase behavior of the baseline kernels. Note that here, the original trajectory is close to linear, hence the linear interpolation happens to behaves artificially well; however, the error in the middle frame shows the subtle departure from non-linearity in the original frames, captured perfectly by our view-aware kernel.

\end{document}